\documentclass[runningheads]{llncs}
\usepackage[T1]{fontenc}
\usepackage{amsfonts}
\usepackage{amssymb}
\usepackage{amsmath} 
\usepackage{graphicx}
\usepackage{booktabs}
\usepackage{multirow}
\usepackage{pifont}
\usepackage[caption=false]{subfig}

\newcommand{\parahead}[1]{\par\medskip\noindent\textbf{#1}\enspace\ignorespaces}

\begin{document}

\title{Do Agents Think Deeper? A Mechanistic Investigation of Layer-Wise Dynamics in Sequential Planning}

\author{Zhenyu Cui \and Xiangzhong Luo\textsuperscript{\textdagger}}
\authorrunning{Z. Cui and X. Luo}
\titlerunning{Do Agents Think Deeper?}
\institute{}
\maketitle              
\begingroup
\renewcommand{\thefootnote}{\textdagger}
\footnotetext{Correspondence: \email{xiangzhong.luo@seu.edu.cn}}
\endgroup
\begin{abstract}
Recent mechanistic studies suggest that large language models (LLMs) may utilize their depth inefficiently in standard single-turn tasks. Whether this still holds in autonomous agent settings, where models must perform multi-turn planning, tool use, and iterative state updates, remains unclear. We study this question through a systematic layer-wise analysis of complete user-agent trajectories spanning three domains: Deep Research, Code Generation, and Tabular Processing. Using residual stream probes, causal layer-skipping interventions, and effective-depth measurements, we show that agentic reasoning exhibits a distinct depth profile from static tasks. As trajectories unfold, models progressively recruit more and deeper layers, with stronger long-range inter-layer dependencies emerging in later turns. At the same time, residual updates become increasingly correction-dominant, indicating a shift from stable feature accumulation toward repeated recalibration. Effective-depth analysis further reveals a substantial construction-refinement gap: semantic direction often forms relatively early, while deep layers remain necessary for stabilizing final outputs. Across model families, this gap is pronounced in Qwen and Minimax, whereas GLM shows a more domain-dependent depth allocation pattern. These results provide mechanistic evidence that autonomous LLM agents allocate depth adaptively as reasoning complexity grows.

\end{abstract}
\section{Introduction}

The rapid evolution of Large Language Models (LLMs)~\cite{openai2024gpt4o,anthropic2024claude,geminiteam2024gemini15unlockingmultimodal,grattafiori2024llama3herdmodels,deepseekai2025deepseekv3technicalreport} has been fundamentally driven by the scaling of parameters, data, and compute. As formalized by foundational scaling laws~\cite{kaplan2020scalinglawsneurallanguage,li2025predictablescaleiifarseer}, predictable performance gains are intrinsically tied to architectural expansion, where model depth emerges as a critical axis. This representational depth enables the hierarchical composition of features, empowering foundation models to sustain multi-step inference and decompose abstract goals into structured plans. These emergent capabilities form the cornerstone of modern autonomous agents~\cite{xi2023risepotentiallargelanguage,Wang_2024}. Unlike static question-answering tasks, agents operate in dynamic environments that demand continuous sequential decision-making. They must select appropriate tools, interpret execution feedback, and maintain a coherent reasoning chain over extended, multi-turn trajectories~\cite{park2023generativeagentsinteractivesimulacra}. Consequently, DeepSeek-R1~\cite{Guo_2025}, OpenAI's o1~\cite{openai2024learning}, and Qwen3~\cite{yang2025qwen3technicalreport} rely heavily on extended computational paths to bridge the gap between static linguistic knowledge and dynamic, executable actions. However, despite the empirical success of deep architectures, a fundamental paradox has recently emerged regarding depth efficiency. As models scale, the performance gains from adding layers begin to diminish, raising a critical question: \textit{Are these foundation models truly leveraging their immense depth to perform complex, hierarchical computations, or are they merely distributing redundant operations across a greater number of layers?}
 
To address this fundamental question, recent mechanistic studies have begun to probe the internal dynamics supporting these capabilities, yielding mixed signals regarding the utility of depth. On one hand, evidence corroborates a reliance on depth for complexity. Investigations into prediction dynamics reveal that complex tasks, such as multi-token fact recall, demand significantly deeper processing layers~\cite{gupta2026llmsusedepth}, and dynamic routing studies show that deeper layers are intensively utilized for iterative refinement in reasoning benchmarks~\cite{heakl2025drllmdynamiclayerrouting}. On the other hand, a provocative study by \cite{hu2025affectseffectivedepthlarge} suggests that in many general NLP tasks, LLMs utilize their depth inefficiently, often spreading computation uniformly or exhibiting redundancy in deeper layers. However, this dichotomy raises a critical question: \textit{Does this inefficient depth hypothesis hold for autonomous agents?} Unlike isolated benchmarks, multi-turn agentic trajectories demand long-horizon state tracking and iterative error correction. We hypothesize that this sequential complexity inherently forces a dynamic and adaptive allocation of depth.

The fundamental limitation of prior mechanistic evaluations lies in their reliance on static, single-turn generations, which inherently strip away this sequential complexity. Constrained by these isolated setups, previous studies naturally concluded that depth is underutilized. However, our preliminary observations in dynamic environments reveal a starkly different phenomenon: as an agent engages in multi-turn tool-use, it appears to actively mobilize deeper computational paths to handle the compounding burden of context maintenance and state updates.

To systematically investigate this contradiction and decode how depth allocation truly evolves, we construct a comprehensive dataset of complete multi-turn trajectories across three challenging domains: Deep Research, Code Generation, and Tabular Processing. By employing Residual Stream Analysis and Layer-Skipping Interventions, our evaluation moves beyond isolated prompts to probe the full reasoning cycle encompassing user queries, internal thoughts, actions, and environmental observations. Tracking how layer efficacy dynamically adapts as the planning horizon expands, this work represents the first systematic attempt to decode the dynamic layer-wise mechanistic behavior of autonomous agents during complex sequential planning. Our main contributions are summarized as follows:

\begin{itemize}
    \item[$\bullet$] \textbf{Discovery of progressive layer mobilization.} We uncover a temporal evolution in depth utilization specific to agentic trajectories. Unlike static tasks, as interaction turns increase, models recruit more layers and establish stronger inter-layer dependencies. This suggests that accumulating trajectory context unlocks deeper computational paths that remain less active in shorter interactions.
    
    \item[$\bullet$] \textbf{Identification of intensified feature refinement cycles.} Through residual stream analysis, we show that internal processing becomes more dynamic as the planning horizon expands. Transitions between feature acquisition and refinement increase in later turns, implying that the agent must continually inject and refine representations to maintain state.
    
    \item[$\bullet$] \textbf{Quantification of a refinement gap in representation maturity.} Using Logit Lens and Effective Depth probes, we identify a clear decoupling between internal feature construction and final output stability. While models often establish semantic direction early, they still require deeper refinement to stabilize output probabilities. This suggests that deep layers are allocated to fine-grained calibration for precise tool execution, unlike the saturation observed in standard tasks.
\end{itemize}

\section{Related Work}

\noindent\textbf{Mechanistic Interpretability of Model Depth.}
While the scaling of Transformer depth has historically been viewed as a primary driver of model capability~\cite{kaplan2020scalinglawsneurallanguage}, enabling the hierarchical composition of features, recent mechanistic interpretations reveal a nuanced picture characterized by significant layer-wise redundancy.
Challenging the strict necessity of sequential depth, \cite{gromov2025unreasonableineffectivenessdeeperlayers} demonstrated the unreasonable ineffectiveness of deeper layers, while \cite{lad2025remarkablerobustnessllmsstages} and \cite{sun2025transformerlayerspainters} observed remarkable robustness to block-wise skipping, suggesting that intermediate layers often perform interchangeable rather than strictly dependent operations.
Corroborating these findings, \cite{gurnee2024languagemodelsrepresentspace} noted that semantic feature construction typically saturates early in the network.
Providing a mechanistic grounding for this inefficiency, \cite{csordás2025languagemodelsusedepth} recently identified a phase transition where the network's latter half is often relegated to iteratively refining output probabilities rather than constructing new features. 
This static depth allocation establishes a strong baseline for standard LLMs, raising the critical question of whether such rigidity persists under the dynamic constraints of agentic workflows.

\medskip

\noindent\textbf{Cognitive Dynamics in Autonomous Agents.}
Unlike static benchmarks that evaluate isolated reasoning steps, autonomous agents operate in dynamic environments necessitating continuous state maintenance and iterative decision-making~\cite{xi2023risepotentiallargelanguage,Wang_2024}.
To support these capabilities, the dominant paradigm has focused on scaling computation along the \textit{temporal} axis, utilizing techniques like Chain-of-Thought (CoT)~\cite{wei2023chainofthoughtpromptingelicitsreasoning} and ReAct~\cite{yao2023reactsynergizingreasoningacting} to externalize intermediate reasoning into the context window.
However, this reliance on reasoning in the token space largely treats the model as a black box, overlooking how internal computational resources are allocated~\cite{mialon2023augmentedlanguagemodelssurvey,liu2023lostmiddlelanguagemodels}.
While recent studies on \textit{latent thinking}~\cite{geiping2025scalingtesttimecomputelatent,hao2025traininglargelanguagemodels} suggest that additional compute improves planning, the specific interplay between extended interaction trajectories and the model's \textit{effective depth} remains underexplored~\cite{packer2024memgptllmsoperatingsystems,csordás2025languagemodelsusedepth}.
Our work bridges this gap by extending the mechanistic analysis of \cite{csordás2025languagemodelsusedepth} into the agentic domain, investigating whether the functional demands of tool use and error correction compel the model to break patterns of shallow saturation and engage in dynamic deepening.

\section{Preliminaries}

In this work, we analyze the internal mechanics of foundation models based on the decoder-only pre-layernorm Transformer architecture~\cite{vaswani2023attentionneed}, within the context of \textit{multi-turn agentic planning}. 
All our analyzed models adhere to the \textbf{Sparse Mixture-of-Experts (MoE)} structure~\cite{shazeer2017outrageouslylargeneuralnetworks,fedus2022switchtransformersscalingtrillion}.

\parahead{Sparse MoE Transformer Architecture.}
An agent trajectory consists of a sequence of turns $\mathcal{T} = \{1, \dots, R\}$. 
At turn $r \in \mathcal{T}$, the model processes a cumulative input sequence $\boldsymbol{x}^{(r)}$, whose length $n_r$ expands progressively (i.e., $n_r > n_{r-1}$). 
The forward pass of layer $l$ at turn $r$ is defined as: 
\begin{align}
    \boldsymbol{a}_l^{(r)} &= \text{SelfAttention}_l(\text{RMSNorm}(\boldsymbol{h}_l^{(r)})) \\
    \hat{\boldsymbol{h}}_l^{(r)} &= \boldsymbol{h}_l^{(r)} + \boldsymbol{a}_l^{(r)} 
\end{align}
Let $\tilde{\boldsymbol{x}}_l^{(r)} = \text{RMSNorm}(\hat{\boldsymbol{h}}_l^{(r)})$ denote the normalized intermediate representation. In a Sparse Mixture-of-Experts (MoE) architecture~\cite{shazeer2017outrageouslylargeneuralnetworks,fedus2022switchtransformersscalingtrillion}, the standard dense Feed-Forward Network is replaced by a routing mechanism that selectively activates a subset of expert networks. The general MoE update is formulated as:
\begin{equation}
    \boldsymbol{m}_l^{(r)} = \sum_{i=1}^{E} g_{l,i}(\tilde{\boldsymbol{x}}_l^{(r)}) \text{FFN}_{l,i}(\tilde{\boldsymbol{x}}_l^{(r)})
\end{equation}
where $E$ is the total number of experts, $\text{FFN}_{l,i}$ is the $i$-th expert network, and $g_{l,i}(\cdot)$ is the routing gate function (typically a sparsely activated Top-$K$ softmax) that computes the routing weight for the $i$-th expert.

Crucially, some advanced architectures analyzed in this work (e.g., GLM-4.5-Air) employ a \textit{Shared-Sparse Expert} paradigm~\cite{dai2024deepseekmoeultimateexpertspecialization}. In this setting, the FFN output is decomposed into a universally activated shared component and dynamically routed sparse components:
\begin{equation}
    \boldsymbol{m}_l^{(r)} = \text{FFN}_{l,\text{shared}}(\tilde{\boldsymbol{x}}_l^{(r)}) + \sum_{i \in \mathcal{S}_K} g_{l,i}(\tilde{\boldsymbol{x}}_l^{(r)}) \text{FFN}_{l,i}(\tilde{\boldsymbol{x}}_l^{(r)})
\end{equation}
where $\text{FFN}_{l,\text{shared}}$ captures universal features across all tokens, and $\mathcal{S}_K$ denotes the set of $K$ dynamically selected sparse experts for the specific token. Finally, the residual stream is updated as:
\begin{equation}
    \boldsymbol{h}_{l+1}^{(r)} = \hat{\boldsymbol{h}}_l^{(r)} + \boldsymbol{m}_l^{(r)}
\end{equation}
Here, $\boldsymbol{h}_l^{(r)} \in \mathbb{R}^{n_r \times d_{\text{model}}}$ is the residual stream~\cite{elhage2021mathematical}, with $\boldsymbol{a}_l^{(r)}$ and $\boldsymbol{m}_l^{(r)}$ representing the Attention and MoE sublayer contributions. RMSNorm~\cite{zhang2019rootmeansquarelayer} is the chosen normalization.

\begin{figure*}[t]
    \centering
    \includegraphics[width=\textwidth]{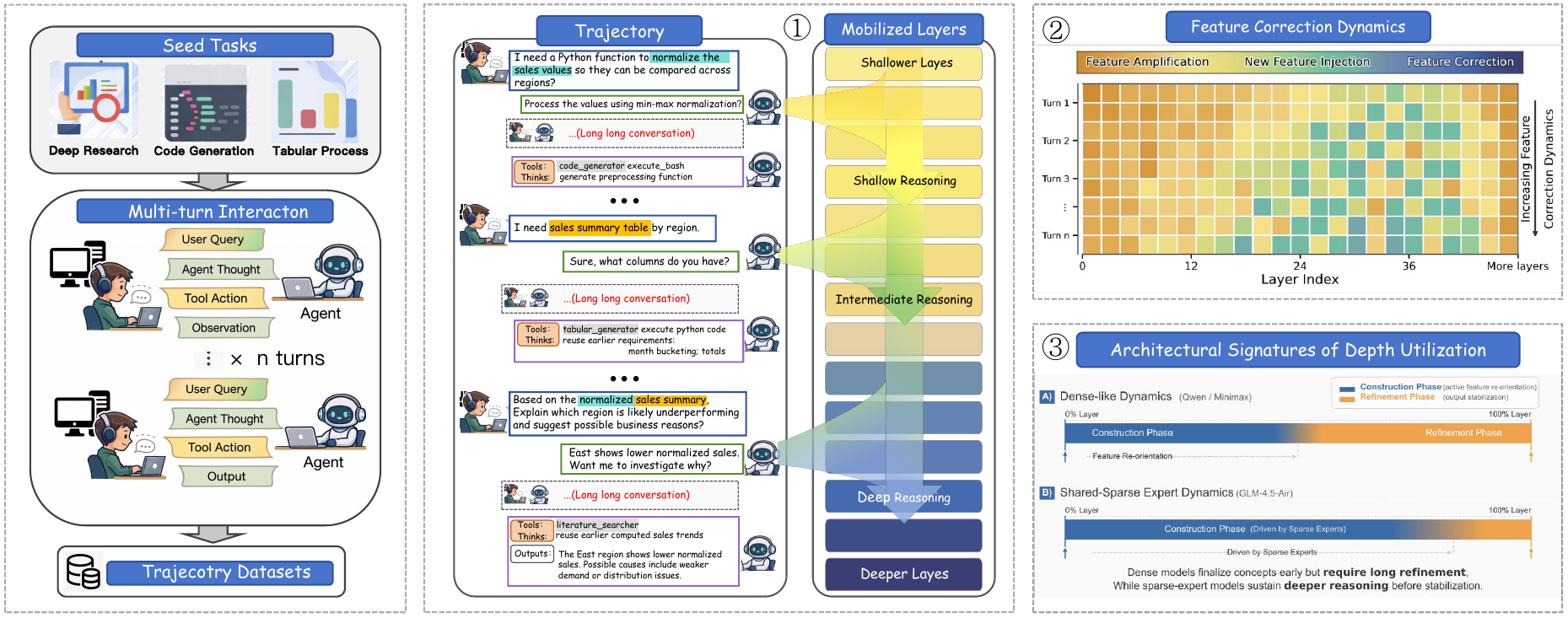}
\caption{
Overview of our study. Left: we construct multi-turn agent trajectories from three seed domains, including Deep Research, Code Generation, and Tabular Processing. Middle: we illustrate a compositional, tool-mediated trajectory in which later turns reuse intermediate artifacts produced earlier (e.g., generated code and derived tables), thereby increasing cross-turn dependency and reasoning complexity. This progression is accompanied by progressively deeper layer mobilization, moving from shallow to intermediate and then deep reasoning. Right: as interaction depth increases, residual updates exhibit more frequent correction dynamics, while model families display distinct depth-utilization signatures, including a construction-refinement gap in Qwen/Minimax and a domain-dependent shared-expert divergence in GLM.
}
\label{fig:overview}
\end{figure*}

\section{Methodology and Analysis}
\vspace{-0.2em}
\parahead{Overview.} To systematically decode the mechanistic behaviors of LLM agents, our investigation is driven by three progressive research questions. First, in \textbf{Sec.~\ref{sec:3.1}}, to determine if agentic planning drives deeper computation, we ask:  {\large{\textcircled{\small{1}}}}\textit{Does the model increasingly rely on deep layers for feature construction as the multi-turn state becomes more complex?} Second, in \textbf{Sec.~\ref{sec:3.2}}, to elucidate how the model qualitatively modifies its internal state, we ask: {\large{\textcircled{\small{2}}}}\textit{Do layer updates transition from simple feature accumulation to active feature correction as the reasoning trajectory extends?} Third, to quantify this depth utilization, \textbf{Sec.~\ref{sec:3.3}} asks:    {\large{\textcircled{\small{3}}}}\textit{When do internal representations reach maturity, and is full depth strictly necessary to stabilize final output probabilities?} Figure~\ref{fig:overview} provides an overview of our framework and key findings, illustrating how multi-turn trajectories reveal progressive layer mobilization, correction-dominant residual dynamics, and architecture-dependent depth utilization patterns.


\vspace{-0.6em} 
\begin{figure*}[t] 
    \centering
    \subfloat[Turn 1\label{fig:sub1}]{%
        \includegraphics[width=0.25\textwidth]{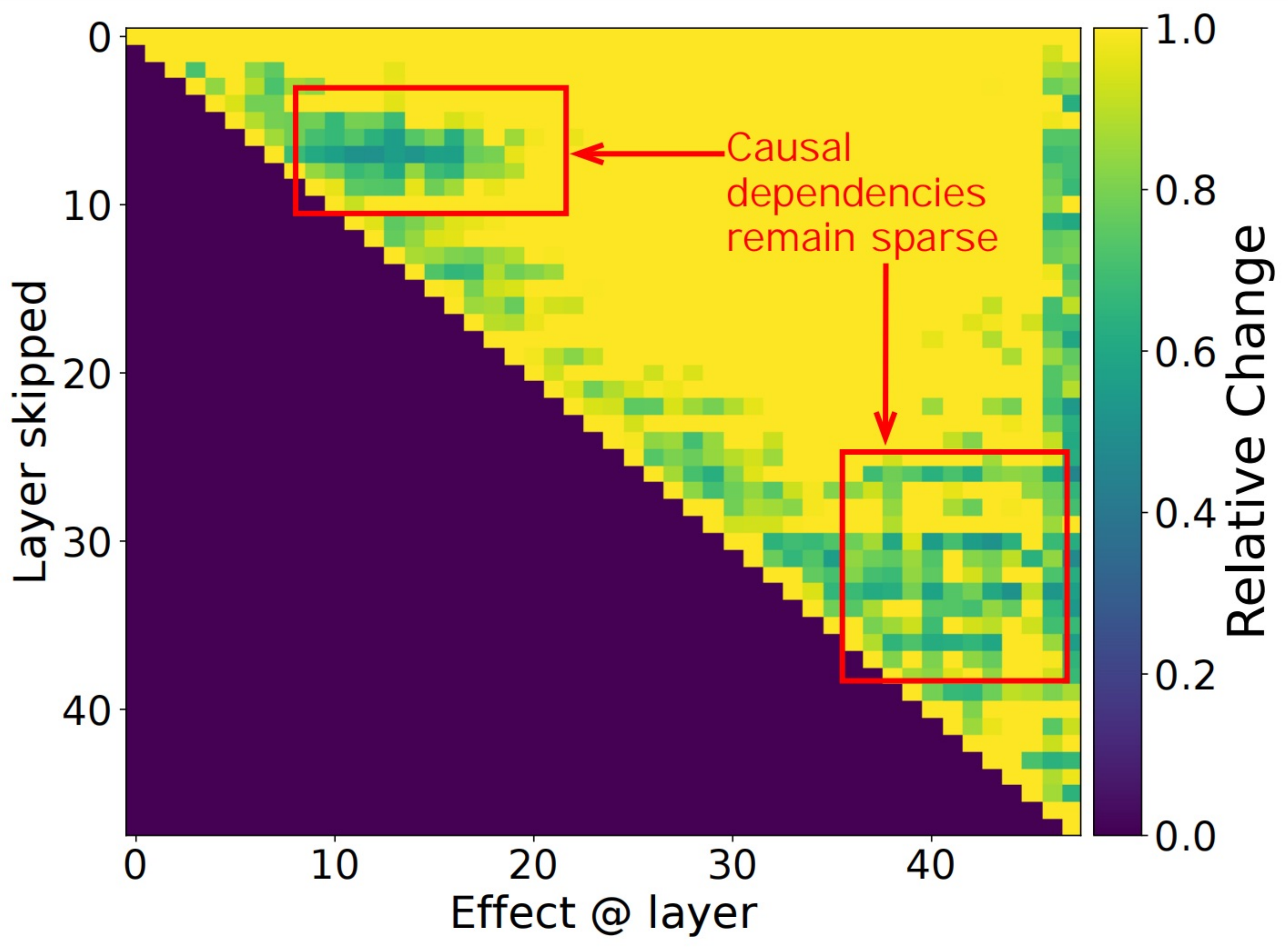}%
    }\hfill 
    \subfloat[Turn 2\label{fig:sub2}]{%
        \includegraphics[width=0.25\textwidth]{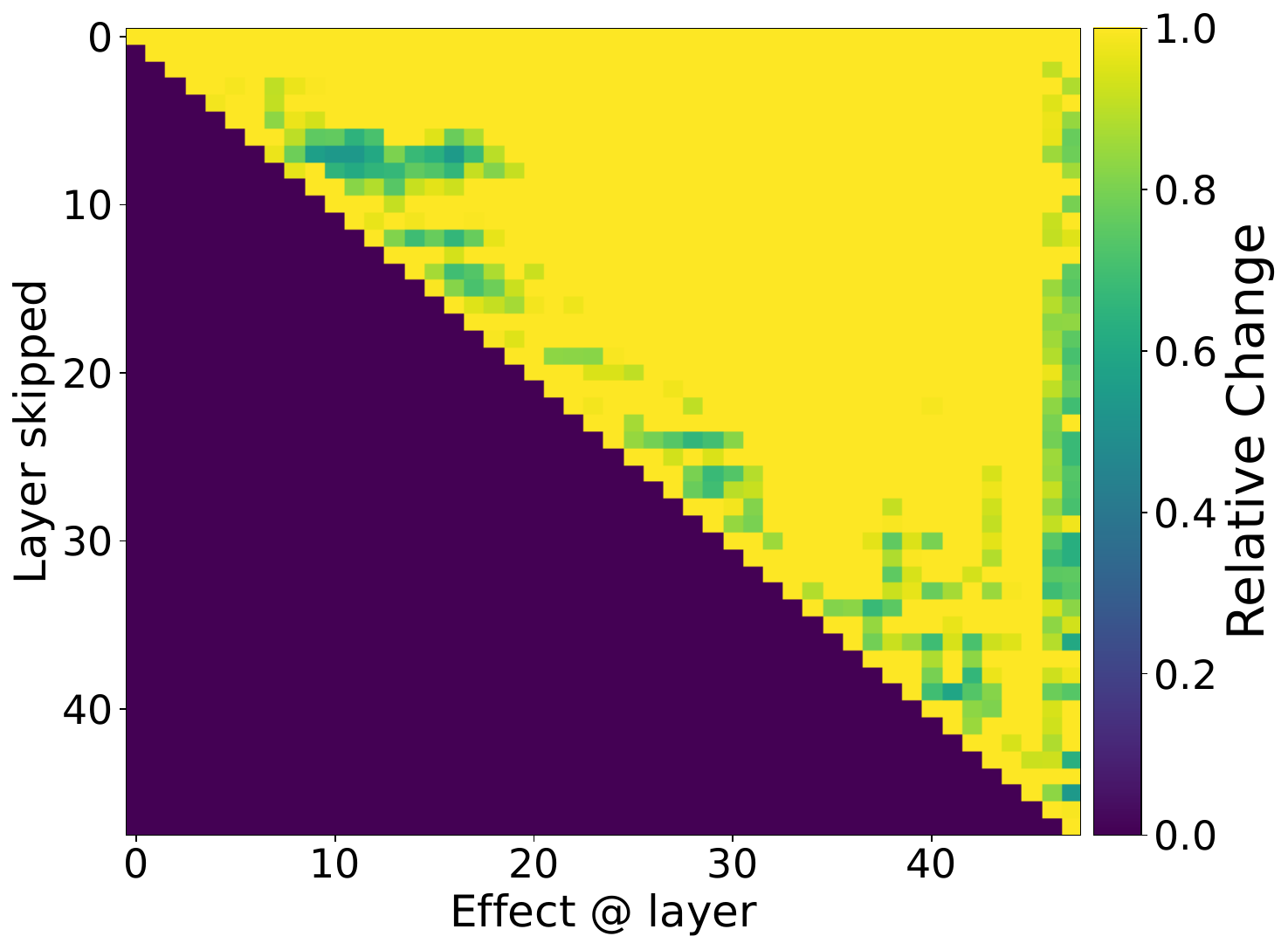}%
    }\hfill
    \subfloat[Turn 3\label{fig:sub3}]{%
        \includegraphics[width=0.25\textwidth]{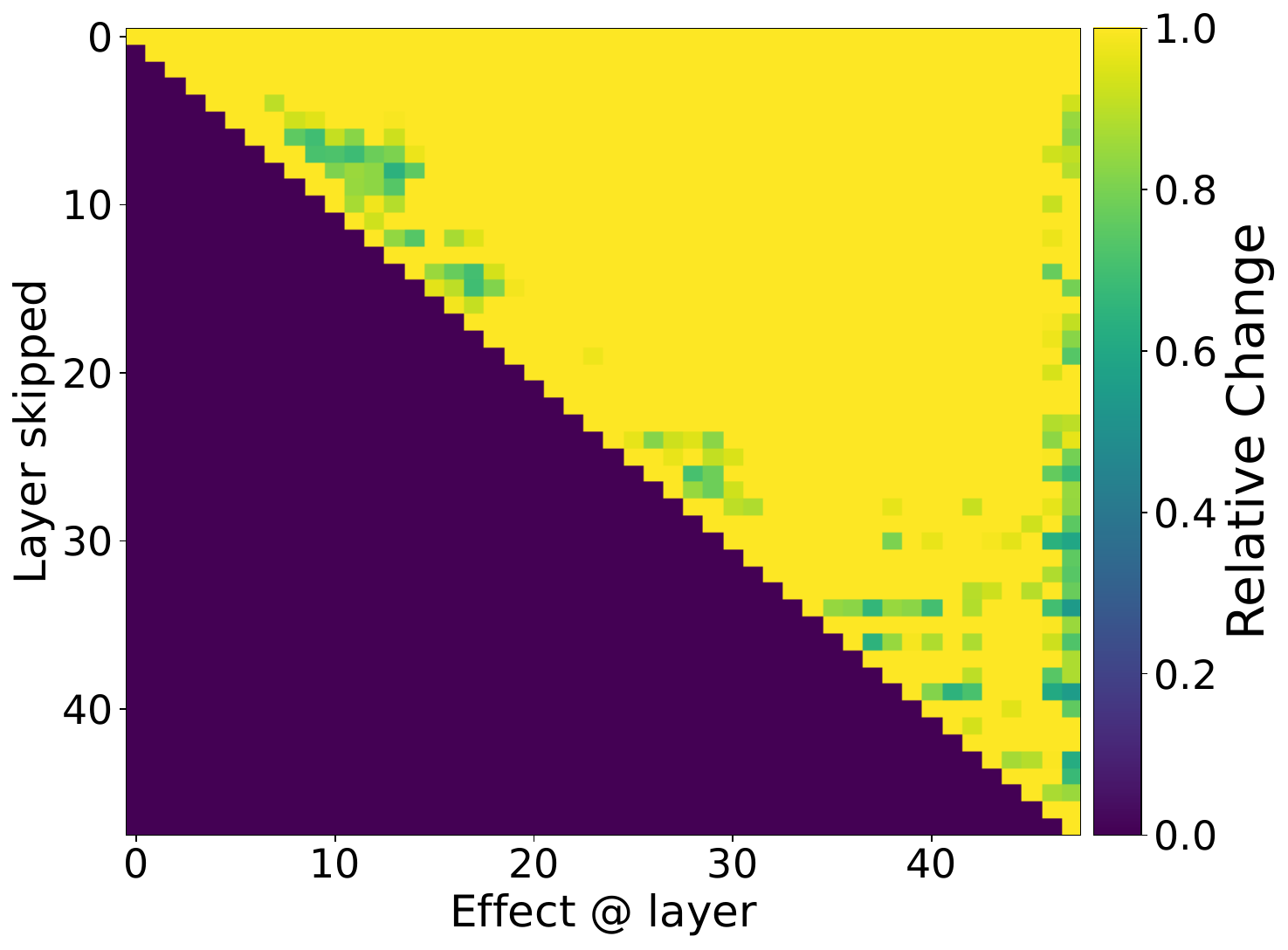}%
    }\hfill
    \subfloat[Turn 4\label{fig:sub4}]{%
        \includegraphics[width=0.25\textwidth]{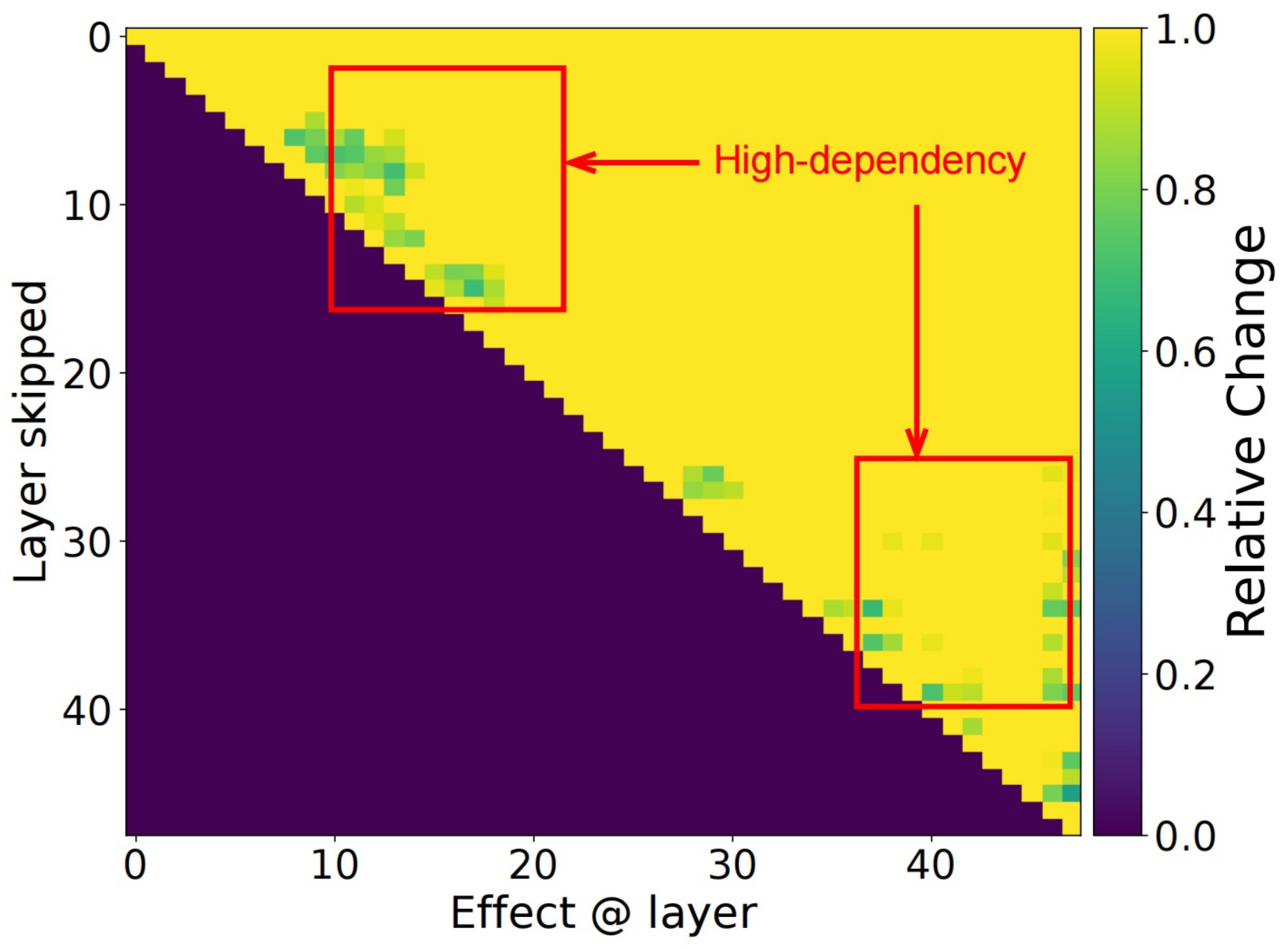}%
    }
    
\caption{
Causal dependency maps across four turns of a representative Code Generation trajectory for Qwen3-Thinking. Each panel shows the Future Effect $E^{(r)}(s,l)$ when an earlier layer $s$ is skipped. Dependencies evolve from sparse, localized interactions in early turns to denser and deeper coupling in later turns.
}

    \label{fig:four_images}
\end{figure*}

\subsection{Dynamic Layer Mobilization via Causal Tracing}
\label{sec:3.1}

\parahead{Setting.}
While various domain-specific benchmarks exist, there is a scarcity of open-source datasets that simultaneously capture the full spectrum of agentic behavior, specifically the interplay of tool use, multi-step planning, and complex reasoning required for a holistic mechanistic analysis. To address this, we innovatively constructed an evaluation suite covering diverse agentic task scenarios. We employ a highly capable agentic model to synthesize complete multi-turn trajectories, explicitly structuring them into interleaved sequences of reasoning (\textit{Thought}), execution (\textit{Action}), and feedback (\textit{Observation}). To ensure validity, seed problems are sourced from established high-quality benchmarks: Deep Research scenarios from the EDR-200  dataset~\cite{prabhakar2025enterprisedeepresearchsteerable}, Code Generation tasks from the SWE-Gym~\cite{pan2025trainingsoftwareengineeringagents}, and Tabular Processing QA from TableLLM ~\cite{zhang2025tablellmenablingtabulardata}.

We begin with a representative case study on Qwen3-30B-Thinking in the Code Generation domain, analyzing a complete multi-turn trajectory to visualize how causal layer dependencies evolve as planning depth increases.

\parahead{Future Effect via Layer Skipping.}
To track how layer mobilization evolves across multiple turns, we quantify the causal dependencies between layers using an intervention method adapted from \cite{csordás2025languagemodelsusedepth}.
For a given turn $r$ and input sequence $\boldsymbol{x}^{(r)}$, we first record the baseline residual stream states $\boldsymbol{h}_l^{(r)}$. 
We then perform an intervention where a specific layer $s$ is effectively skipped for all tokens starting from a position $p$. This is implemented by setting the intervened state $\tilde{\boldsymbol{h}}_{s+1}^{(r)}[p:] = \tilde{\boldsymbol{h}}_s^{(r)}[p:]$, thereby nullifying the contribution of layer $s$ for the subsequent sequence.

We measure the impact of this ablation on a downstream layer $l > s$ by computing the maximum relative change in its contribution vector.
Formally, the \textit{Future Effect} metric $E^{(r)}(s, l)$ is defined as:

\begin{equation}
    E^{(r)}(s, l) = \max_{p} \frac{\| \boldsymbol{C}_l^{(r)} - \tilde{\boldsymbol{C}}_l^{(r)} \|_2}{\| \boldsymbol{C}_l^{(r)} \|_2}
\end{equation}

where $\boldsymbol{C}_l^{(r)} = \boldsymbol{h}_{l+1}^{(r)} - \boldsymbol{h}_l^{(r)}$ is the original contribution of layer $l$ at turn $r$, and $\tilde{\boldsymbol{C}}_l^{(r)}$ is the contribution under the intervention.
This metric captures the causal necessity of layer $s$: a high value implies that layer $l$'s computation critically depends on the features generated by layer $s$. We select the maximum value over sequence positions $p$ to capture sparse but critical reasoning steps typical in agentic logic.

\begin{figure*}[t] 
    \centering
    \subfloat[Turn 1\label{fig:logit_turn1}]{%
        \includegraphics[width=0.25\textwidth]{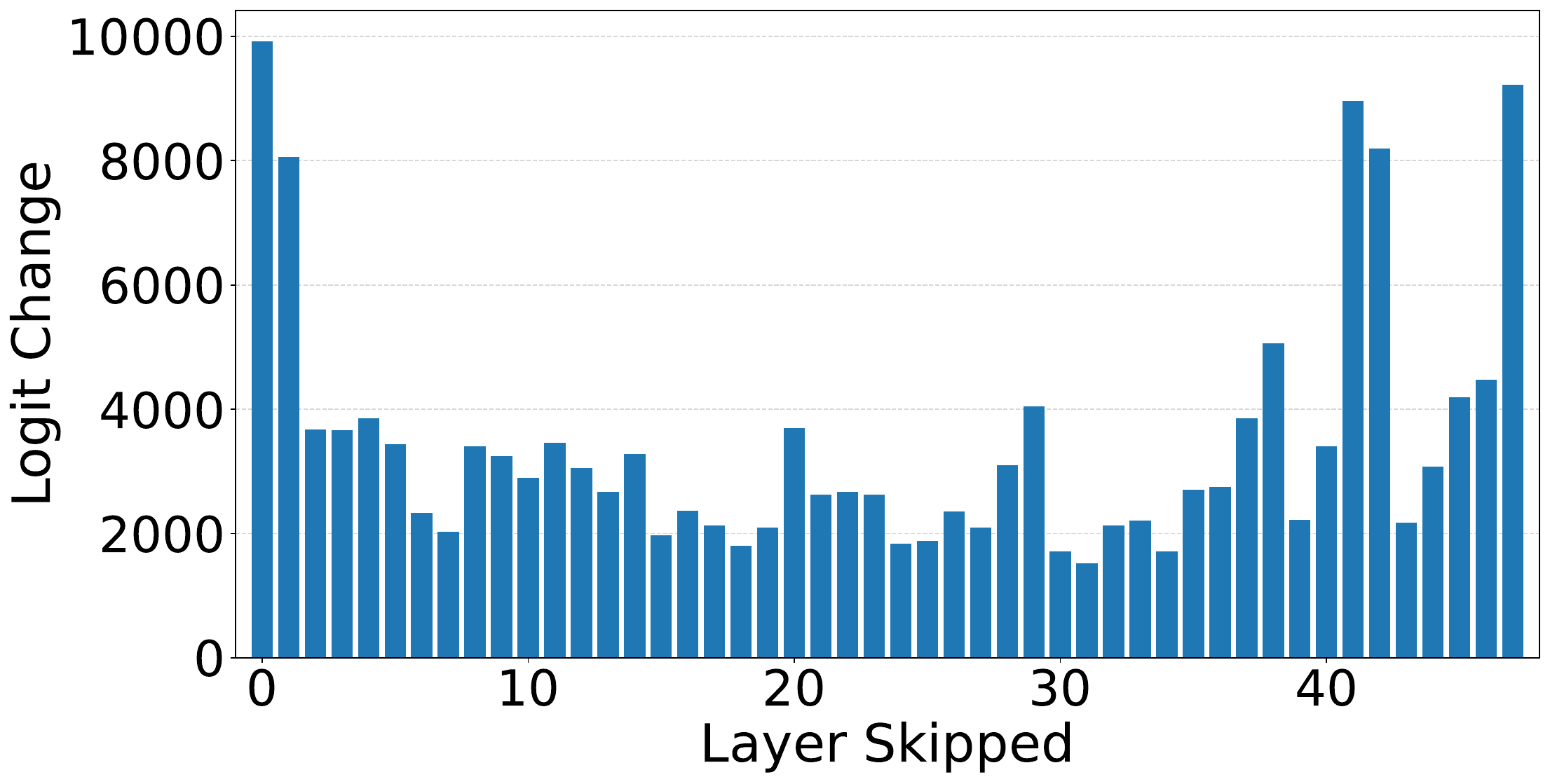}%
    }\hfill 
    \subfloat[Turn 2\label{fig:logit_turn2}]{%
        \includegraphics[width=0.25\textwidth]{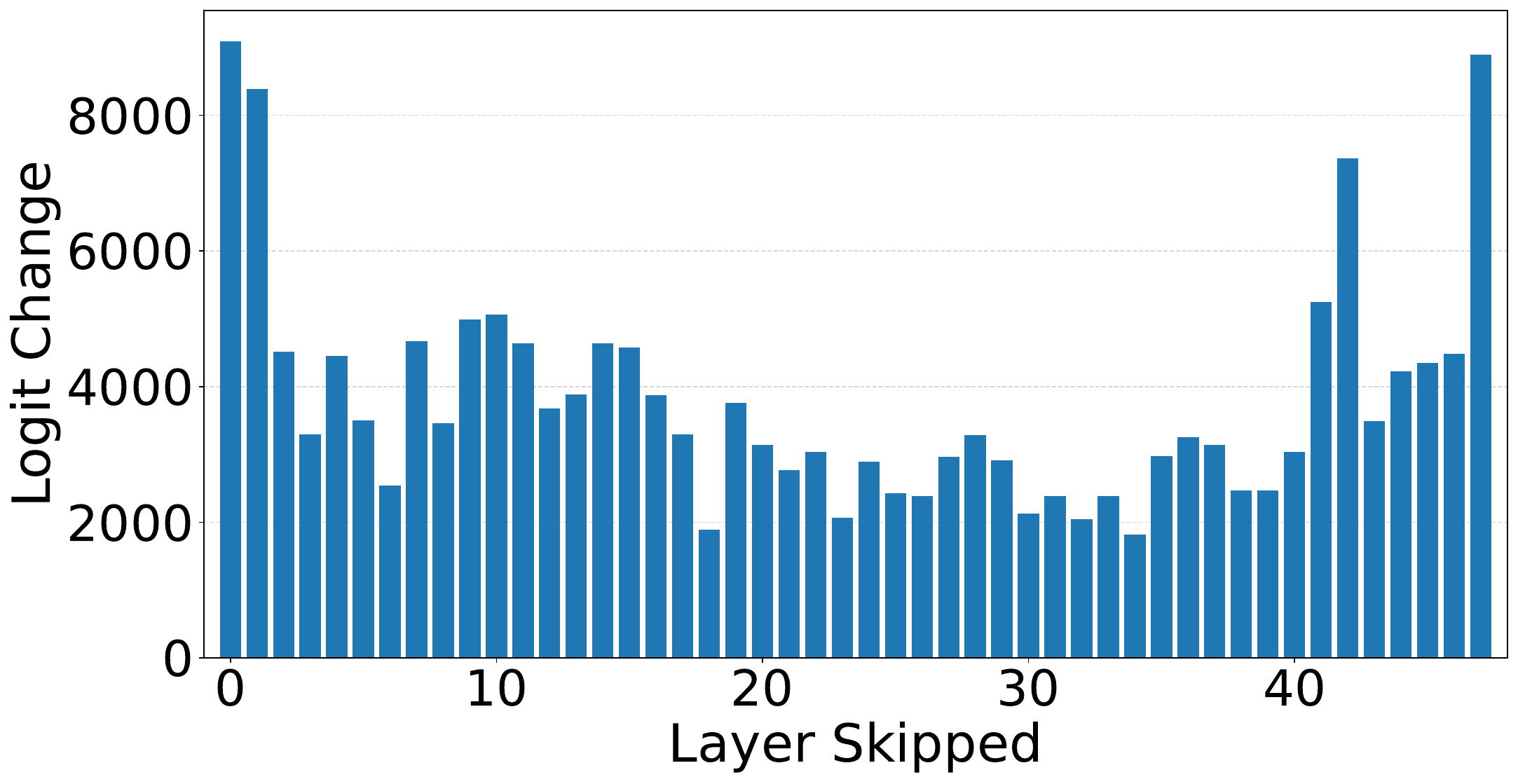}%
    }\hfill
    \subfloat[Turn 3\label{fig:logit_turn3}]{%
        \includegraphics[width=0.25\textwidth]{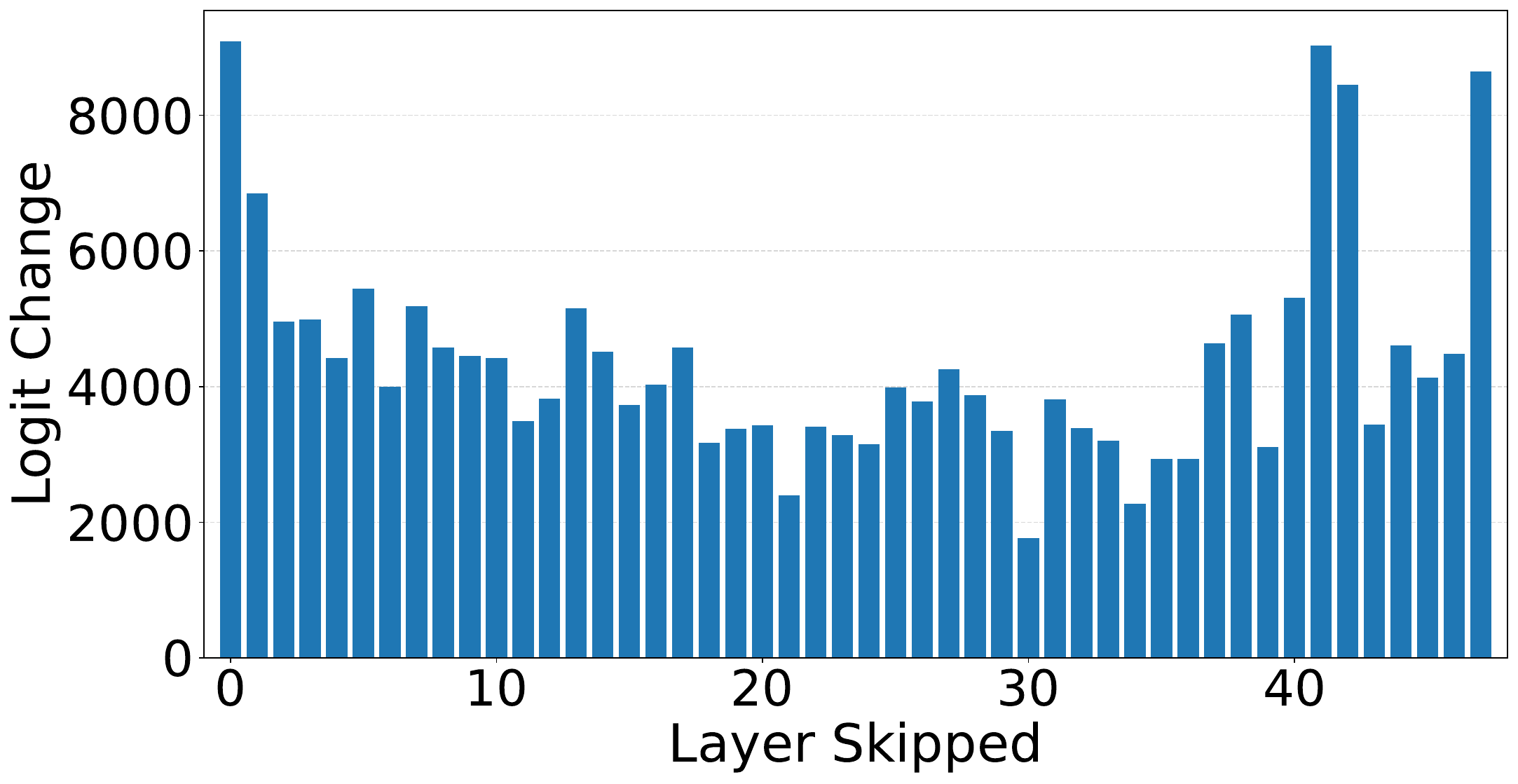}%
    }\hfill
    \subfloat[Turn 4\label{fig:logit_turn4}]{%
        \includegraphics[width=0.25\textwidth]{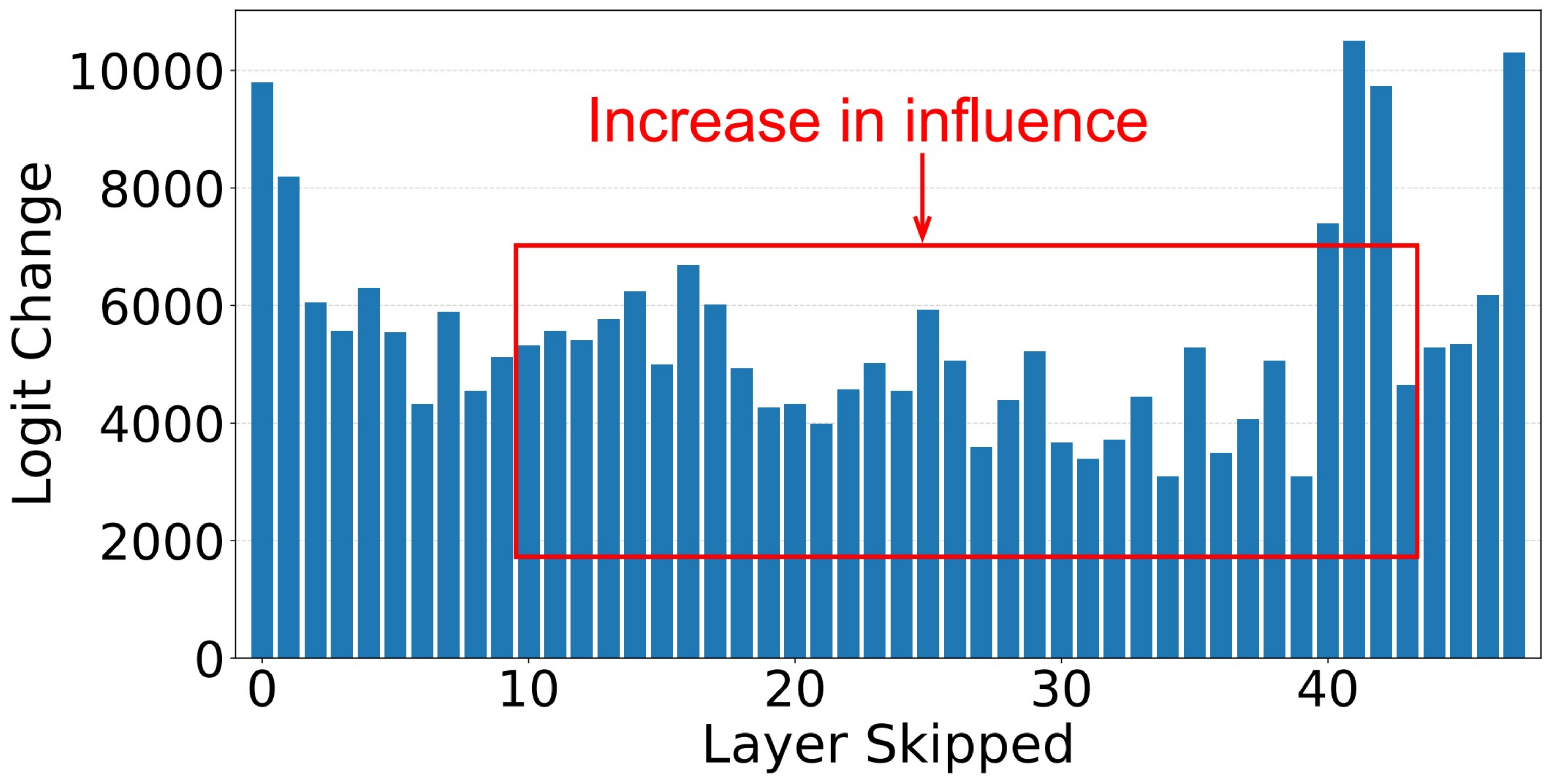}%
    }
    
\caption{
Layer-wise impact on future predictions across four turns of the same Code Generation trajectory (Qwen3-Thinking). Bar height indicates the Logit Change Norm $D^{(r)}(s)$. As the interaction progresses, future predictions depend on a broader and deeper set of layers, indicating progressively greater depth utilization in later turns.
}
    \label{fig:logit_diffs}
\end{figure*}

\medskip

\parahead{Progressive Layer Mobilization.}
Figure~\ref{fig:four_images} visualizes the model's internal topology, displaying the evolution of causal dependencies across a complete multi-turn task. From Turn 1 to Turn 4, causal dependencies become markedly denser and deeper. In the initial turn (Fig.~\ref{fig:sub1}), causal dependencies remain sparse and highly localized. However, as the agent progresses to later turns (Fig.~\ref{fig:sub4}), which require more complex state updates and reasoning, we observe a striking expansion of high-dependency structural linkages. This transition from isolated local effects to dense, long-range causal chains confirms that, as the interaction context deepens, the model increasingly mobilizes its full depth, rendering earlier-layer features indispensable for downstream computations.

To quantify the impact on \textit{future} token generation, we sample a pivot position $t_s$, intervene on layer $l$ for context tokens $t \leq t_s$, and measure the predictive divergence for the subsequent sequence $t > t_s$. The results reveal a dramatic turn-dependent shift. In the initial turn (Fig.~\ref{fig:logit_turn1}), our observations align with static task baselines: the intermediate-to-deep layers exert minimal influence on future computations, exhibiting a classic U-shaped importance profile. Crucially, however, this pattern is fundamentally disrupted as the agent enters deeper planning stages (Fig.~\ref{fig:logit_turn4}). Here, the previously dormant intermediate-to-deep layers (e.g., Layers 10-45) exhibit a pronounced surge in influence, confirming their necessity for sustained multi-turn execution.

\medskip

\begin{figure*}[t]
    \centering
    \subfloat[Set A: Turn 1]{%
        \includegraphics[width=0.25\textwidth]{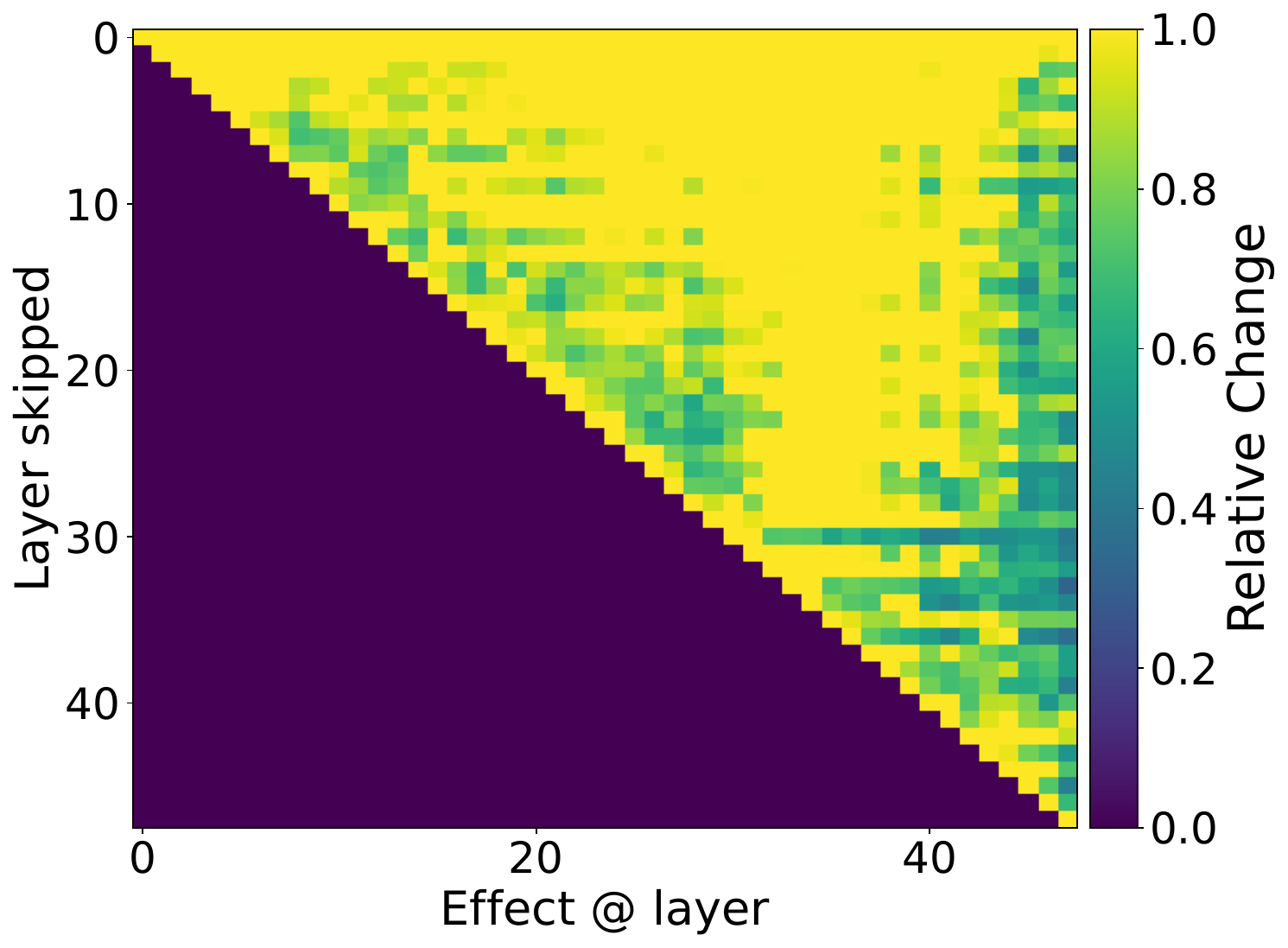}%
    }\hfill
    \subfloat[Set A: Turn 4]{%
        \includegraphics[width=0.25\textwidth]{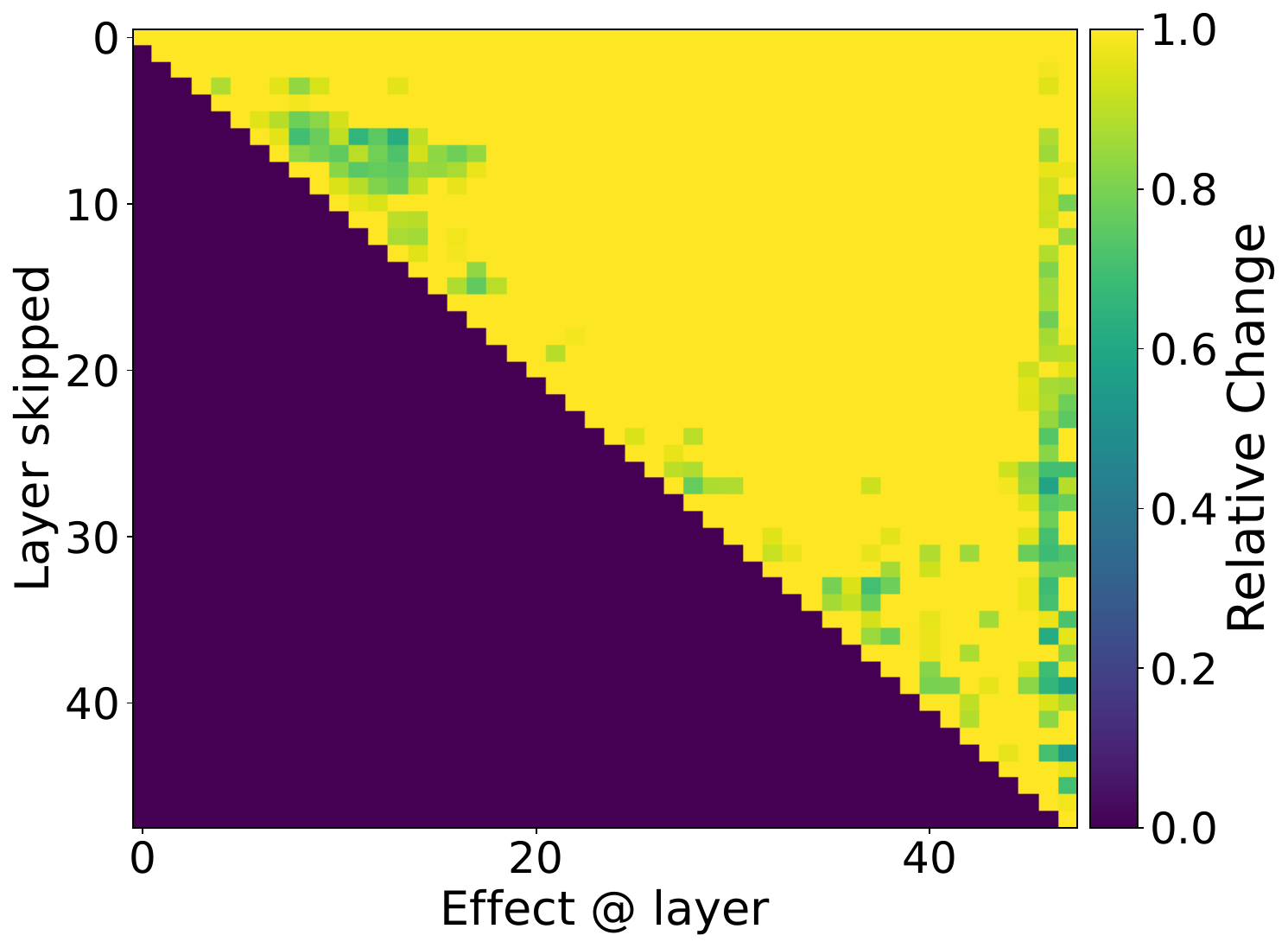}%
    }\hfill
    \subfloat[Set B: Turn 1]{%
        \includegraphics[width=0.25\textwidth]{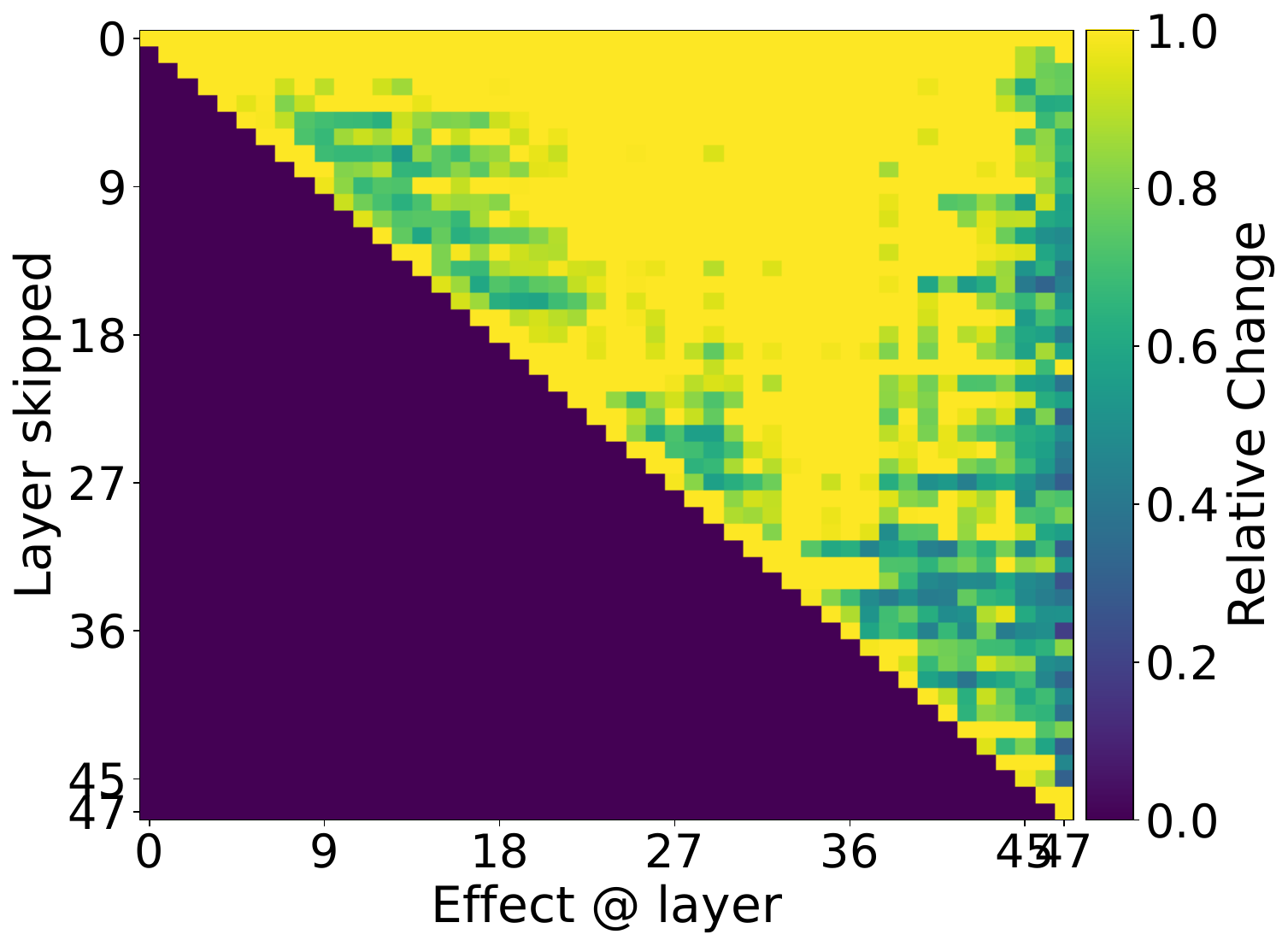}%
    }\hfill
    \subfloat[Set B: Turn 4]{%
        \includegraphics[width=0.25\textwidth]{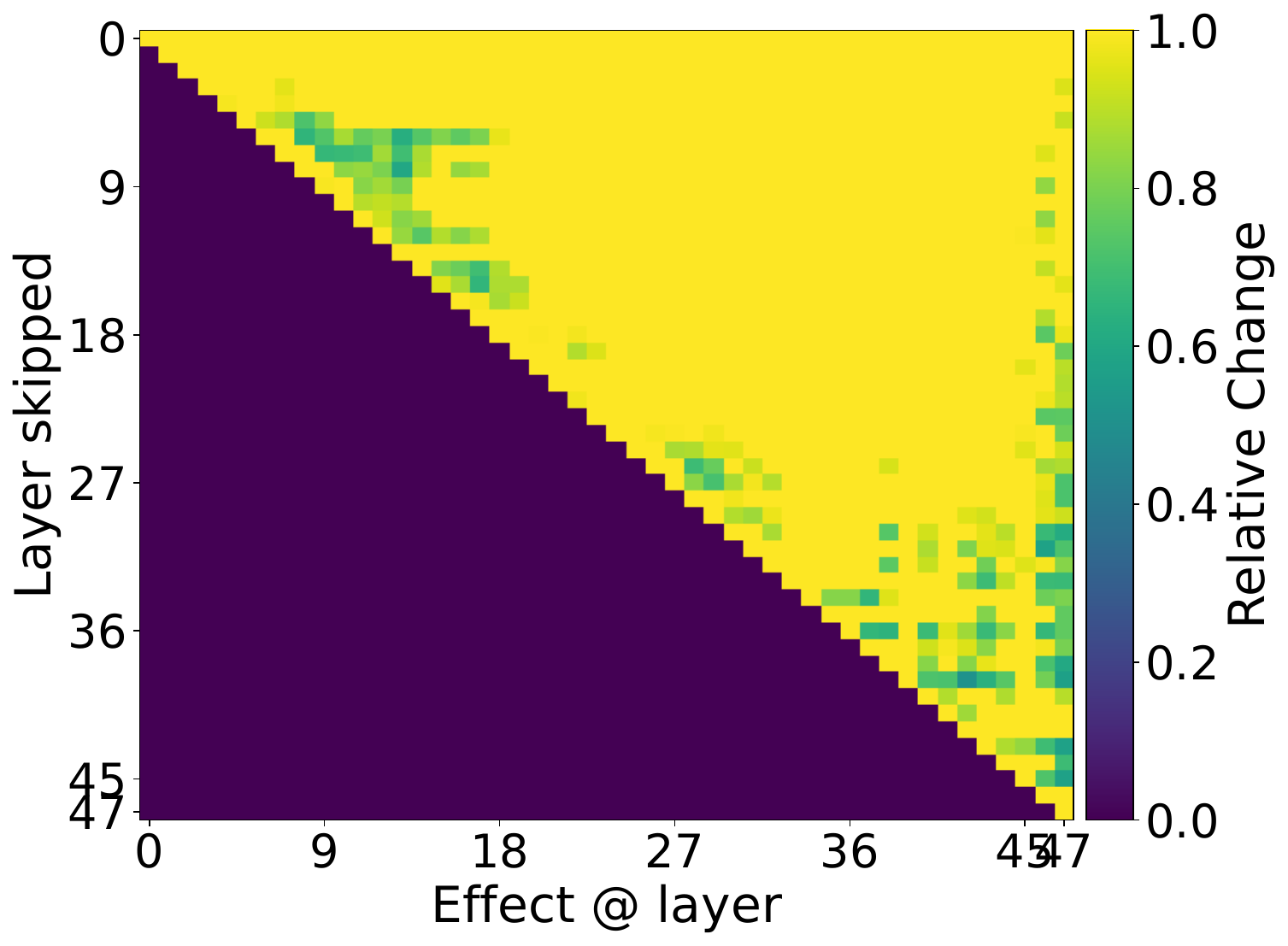}%
    }

\caption{
Intra-task consistency check on two non-overlapping validation subsets. Both subsets show the same transition from relatively localized dependencies in Turn 1 to broad full-depth mobilization in Turn 4.
}
    \label{fig:sample_consistency}
\end{figure*}

\parahead{Validating the Universality of Dynamic Deepening.} To verify that the observed dynamic layer mobilization is not an artifact of sample selection, we repeat the causal tracing analysis on two non-overlapping validation subsets (A and B) drawn from the Deep Research, Code Generation, and Tabular Processing domains. As shown in Fig.~\ref{fig:sample_consistency}, both subsets exhibit the same transition: dependencies are relatively localized in the initial turn, but expand to near full-depth structures in later turns. This suggests that progressive depth allocation reflects a systematic response to increasing reasoning complexity rather than prompt-specific variation.

\begin{figure*}[t]
    \centering
    \subfloat[Minimax: Turn 1]{%
        \includegraphics[width=0.25\textwidth]{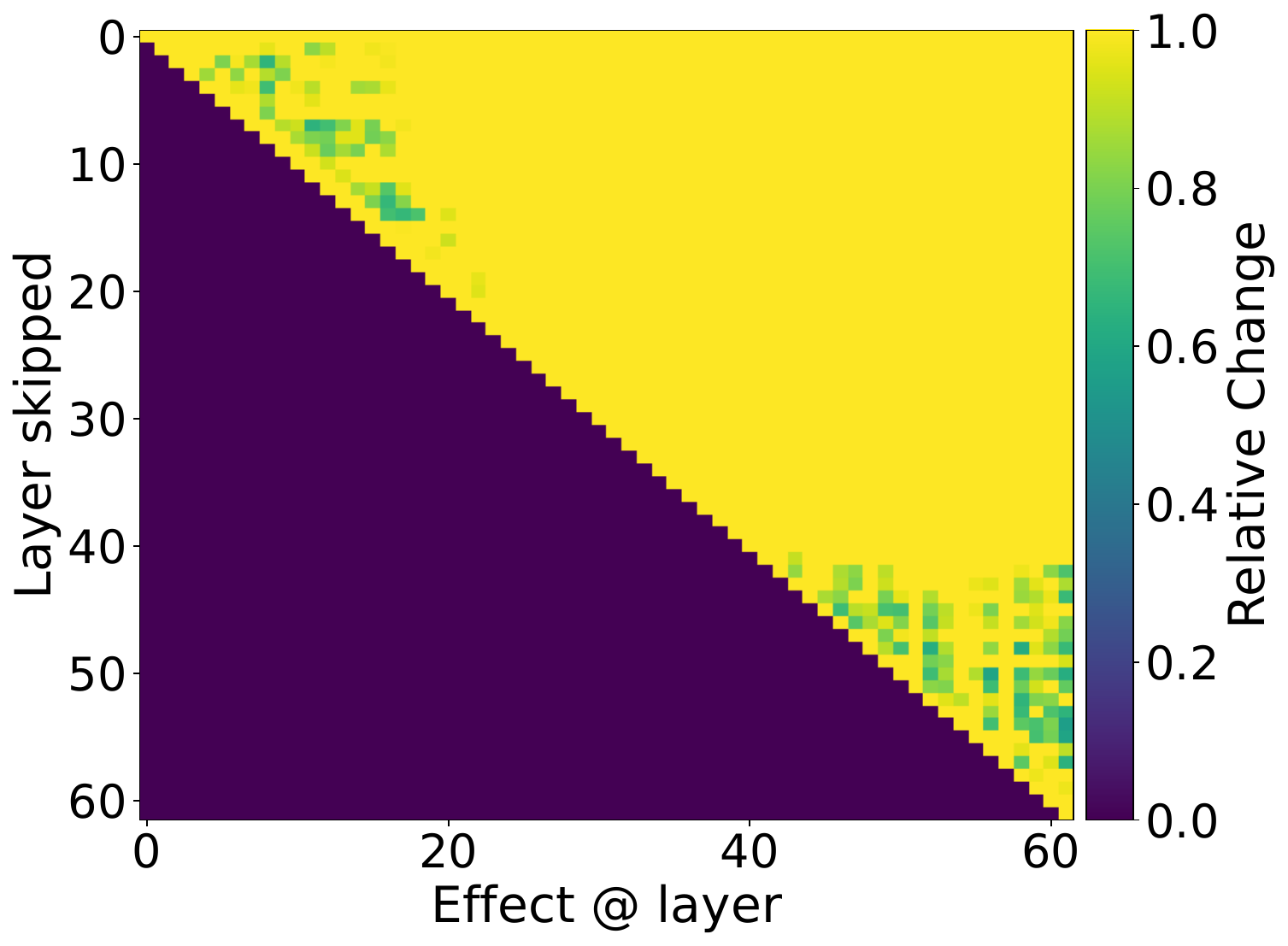}%
    }\hfill
    \subfloat[Minimax: Turn 4]{%
        \includegraphics[width=0.25\textwidth]{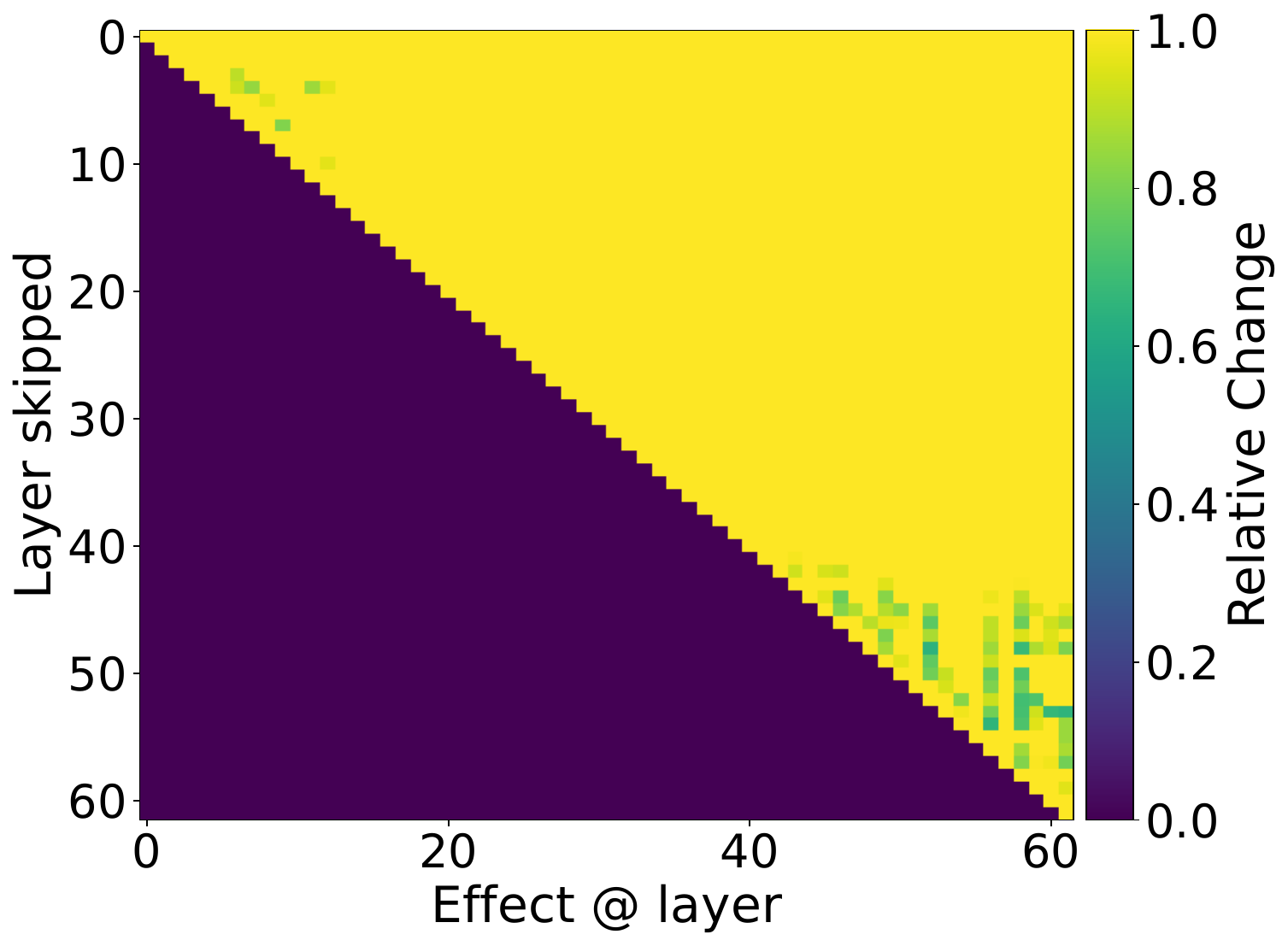}%
    }\hfill
    \subfloat[Qwen: Turn 1]{%
        \includegraphics[width=0.25\textwidth]{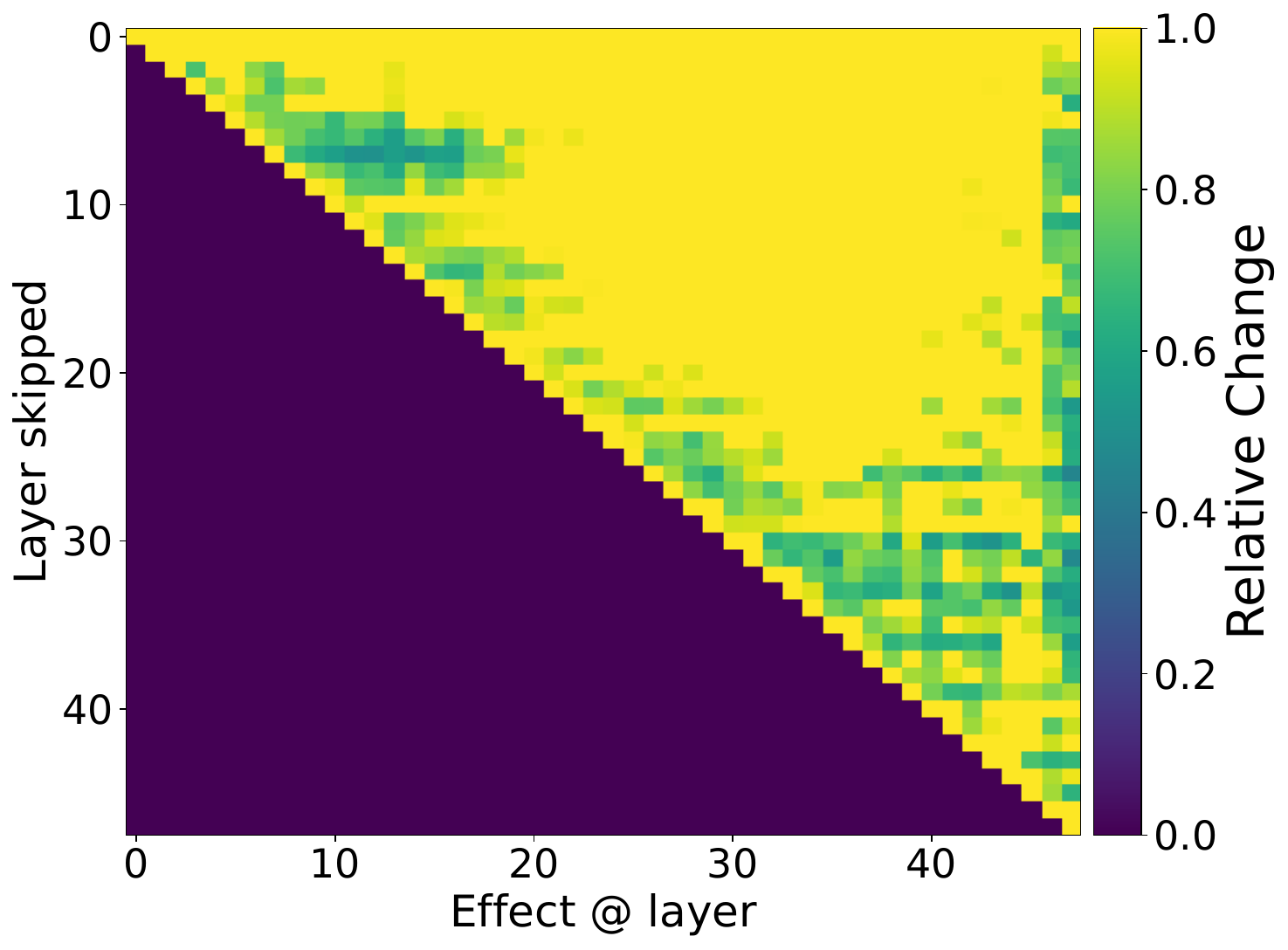}%
    }\hfill
    \subfloat[Qwen: Turn 4]{%
        \includegraphics[width=0.25\textwidth]{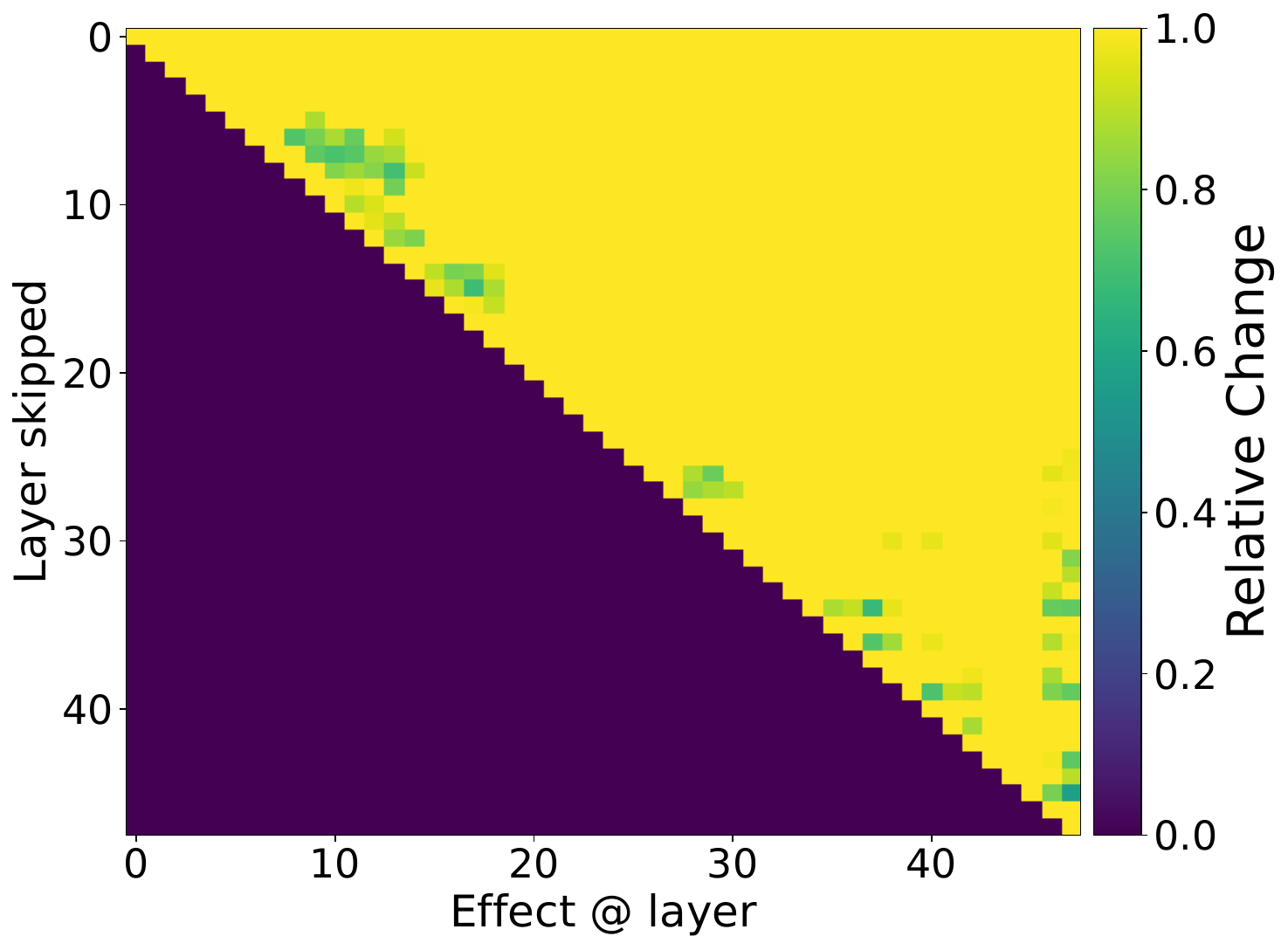}%
    }

\caption{
Cross-model consistency check under synchronous (Minimax on Minimax-generated trajectories) and asynchronous (Qwen on Minimax-generated trajectories) evaluation. Both settings exhibit the same shift from sparse early-turn dependencies to broad full-depth mobilization in later turns.
}
    \label{fig:cross_model_robustness}
\end{figure*}

A second potential confounder is the alignment between the trajectory generator and the evaluator. Because the trajectories are generated by Minimax, the observed deepening could reflect a synchronous setting in which the model processes its own stylistic outputs. To control for this factor, we apply the same causal analysis to Qwen3 on the same Minimax-generated trajectories. Figure~\ref{fig:cross_model_robustness} shows that both synchronous and asynchronous settings exhibit the same shift, from sparse early-turn dependencies to broad full-depth mobilization in later turns. This suggests that dynamic depth allocation is driven primarily by the semantic complexity of the evolving task state rather than model-specific stylistic alignment.

\subsection{Dynamic Feature Evolution via Residual Cosine Similarity}
\label{sec:3.2}
While Sec.~\ref{sec:3.1} reveals \textit{where} critical computation occurs, it does not elucidate \textit{how} the model modifies its internal state. To determine whether extended reasoning shifts internal processing from simple feature accumulation toward active feature correction, we analyze the geometric alignment between layer updates and the residual stream.

\parahead{Setting.}
To examine how internal feature updates evolve over time, we compare residual cosine similarity patterns for Qwen3-30B-Instruct and Qwen3-30B-Thinking between early and late turns on matched agentic trajectories, using the same setup as in Sec.~\ref{sec:3.1}.

\parahead{Residual Cosine Similarity.}
Let $\boldsymbol{u}_l^{(r)} = \boldsymbol{a}_l^{(r)} + \boldsymbol{m}_l^{(r)}$ denote the total additive contribution of layer $l$ at turn $r$. We quantify the alignment between this update and the incoming residual stream $\boldsymbol{h}_l^{(r)}$ as:
\begin{equation}
    S_{\text{layer}}^{(r)}(l) = \text{cossim}(\boldsymbol{u}_l^{(r)}, \boldsymbol{h}_l^{(r)})
\end{equation}
where $\text{cossim}(\mathbf{x}, \mathbf{y}) = \frac{\mathbf{x} \cdot \mathbf{y}}{\|\mathbf{x}\|_2 \|\mathbf{y}\|_2}$.

\begin{figure*}[t]
    \centering
    \subfloat[Qwen3-Instruct (Turn 1)\label{fig:ins_r1}]{%
        \includegraphics[width=0.48\textwidth]{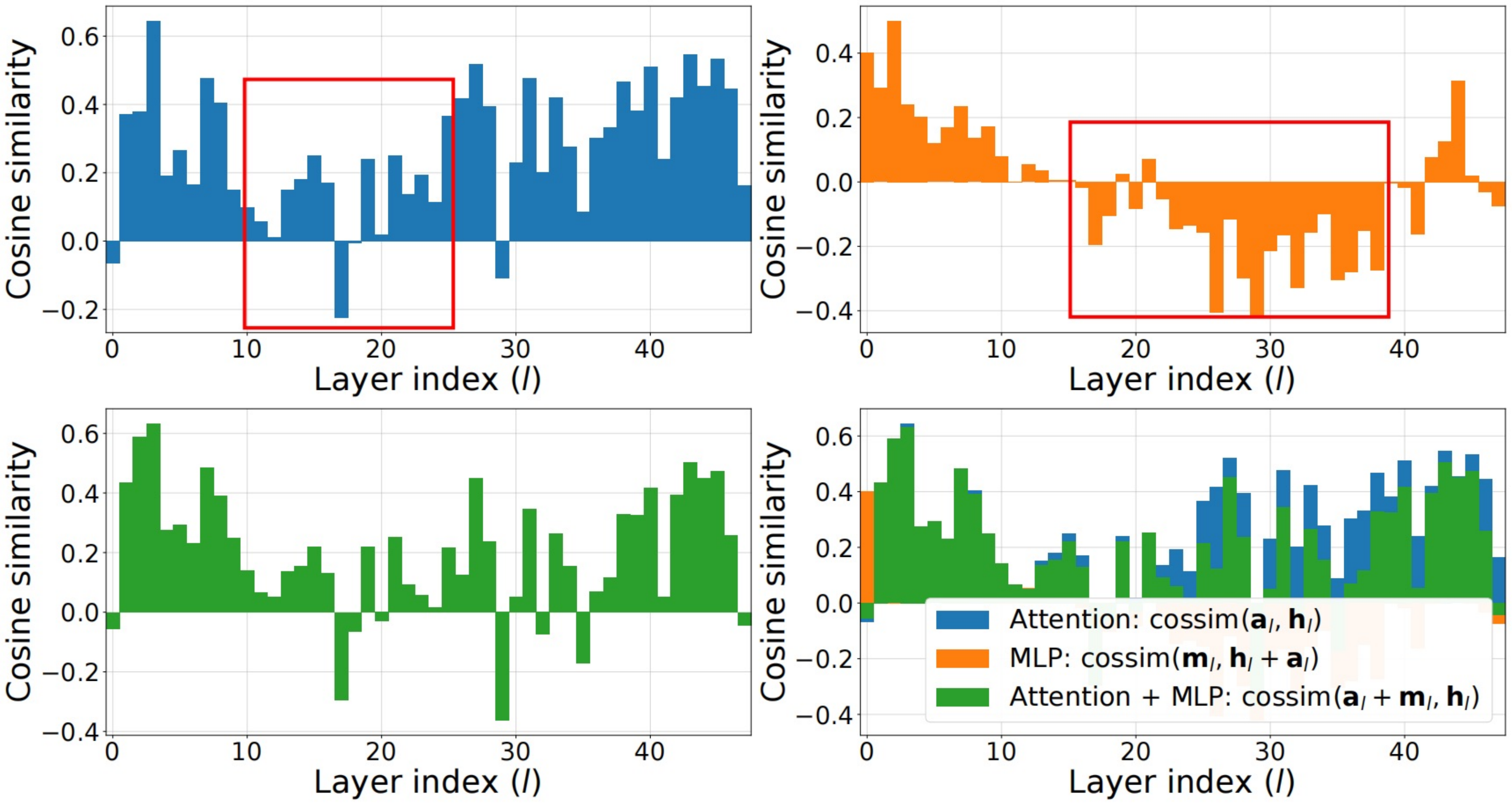}%
    }\hfill
    \subfloat[Qwen3-Instruct (Turn 5)\label{fig:ins_r5}]{%
        \includegraphics[width=0.48\textwidth]{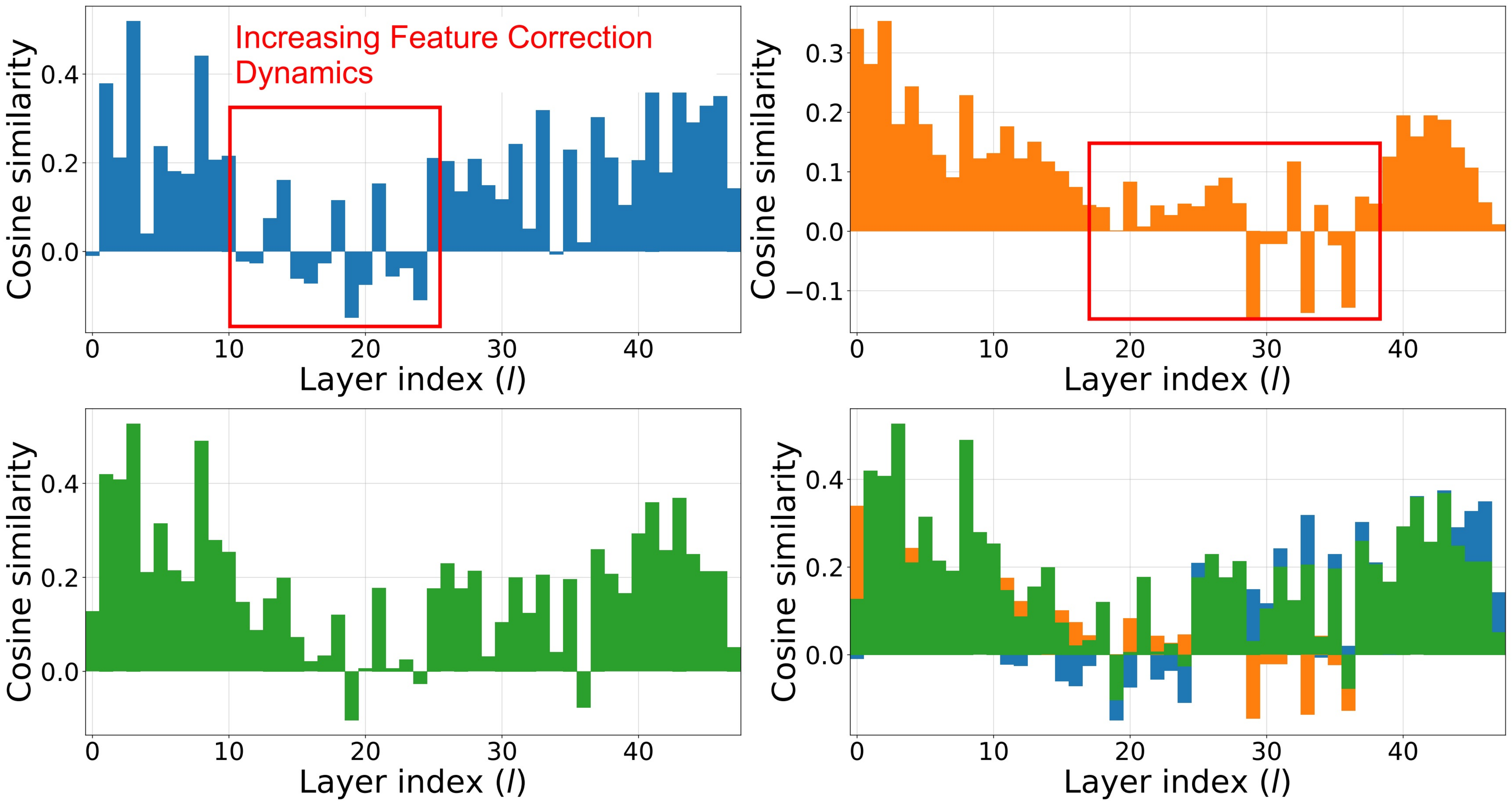}%
    }
    
    \vspace{10pt} 

    \subfloat[Qwen3-Thinking (Turn 1)\label{fig:think_r1}]{%
        \includegraphics[width=0.48\textwidth]{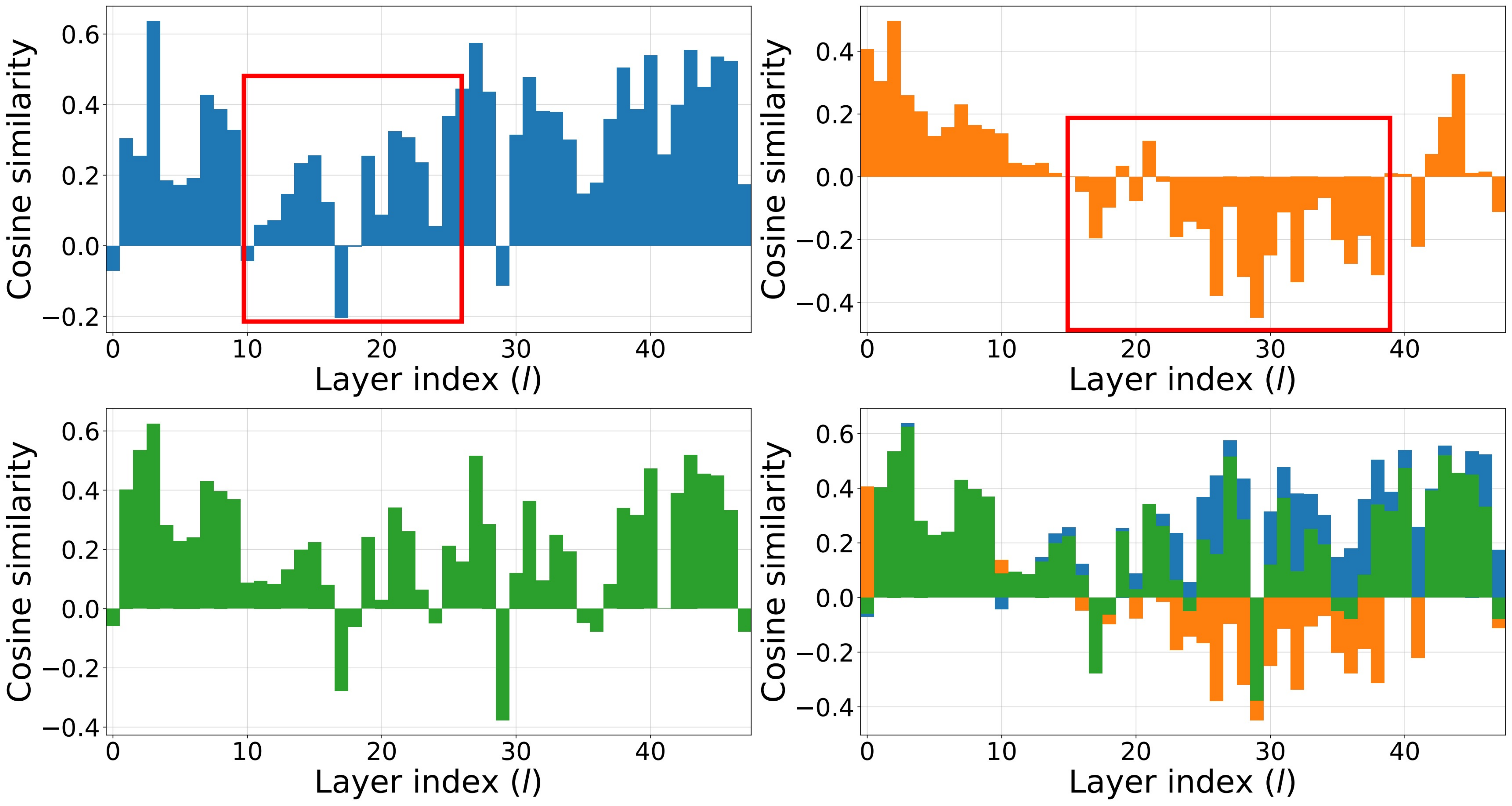}
    }\hfill
    \subfloat[Qwen3-Thinking (Turn 5)\label{fig:think_r5}]{%
        \includegraphics[width=0.48\textwidth]{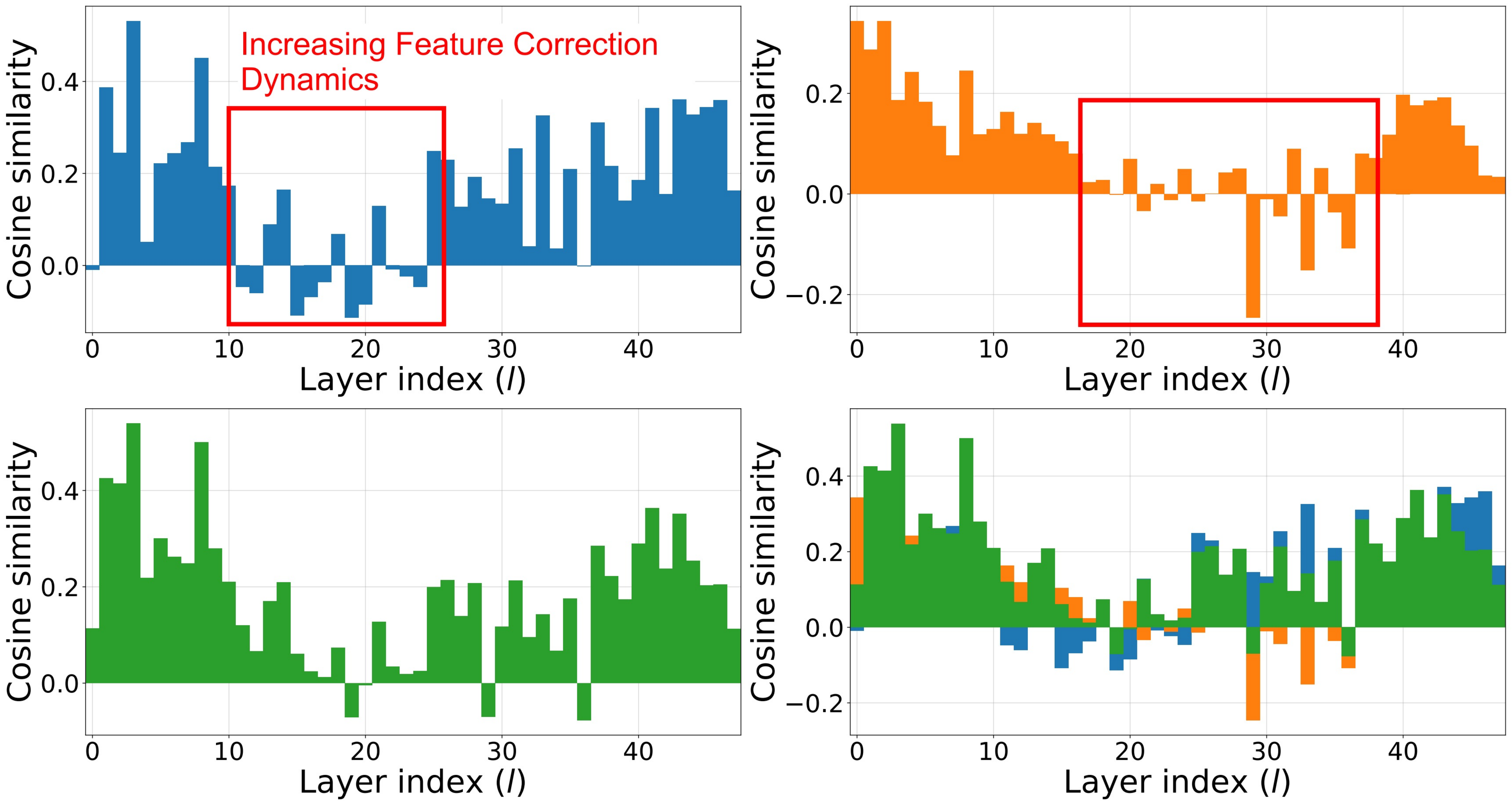}%
    }
    \caption{
Residual cosine similarity patterns for Qwen3-Instruct (top) and Qwen3-Thinking (bottom), comparing Turn 1 (left) and Turn 5 (right). Later turns exhibit more frequent phase changes, especially in intermediate and deep layers, indicating a shift from stable feature amplification toward more active feature correction.
}
    \label{fig:cosine_similarity_matrix}
\end{figure*}

\parahead{Mechanistic Interpretation of Alignment.}
Following \cite{csordás2025languagemodelsusedepth}, we interpret $S_{\text{layer}}^{(r)}(l)$ as a proxy for the functional role of layer updates, with three regimes:
\begin{itemize}
    \renewcommand\labelitemi{$\bullet$}
    \item \textbf{Orthogonal injection ($S \approx 0$):} the layer introduces features largely orthogonal to the current state, suggesting new information injection.
    \item \textbf{Feature amplification ($S > 0$):} the update aligns with the residual direction, reinforcing existing semantic features.
    \item \textbf{Feature correction ($S < 0$):} the update opposes the residual direction, indicating suppression or redirection of previously accumulated features.
\end{itemize}

\parahead{Intensification of Feature Correction Dynamics.}
Figure~\ref{fig:cosine_similarity_matrix} shows a clear temporal shift in these alignment patterns. Early turns are dominated by relatively stable positive alignment, consistent with gradual feature accumulation. As the trajectory lengthens, however, the alignment profiles become increasingly oscillatory, with frequent sign changes emerging in intermediate and deep layers.

As the interaction progresses from Turn 1 to Turn 5, residual stream dynamics exhibit a marked increase in \textbf{phase changes}. In the Attention updates, initially smooth alignment patterns break into repeated positive-negative transitions; in the MoE updates, broad amplification regions observed in Turn 1 give way to alternating correction-amplification regimes by Turn 5. The result is a substantially more dynamic residual evolution in later turns.

This collective behavior suggests that later turns mobilize deeper layers to actively re-orient the internal state, contrasting sharply with the monotonic transformations of initial queries. Consequently, \textbf{high-frequency feature correction emerges as a fundamental signature of deep iterative planning}. As context accumulates, simple feature addition becomes insufficient; the model must repeatedly correct and calibrate the information flow.

\subsection{Probing Representation Maturity and Depth Necessity}
\label{sec:3.3}

While the preceding analyses reveal dynamic layer mobilization, they do not specify \textit{when} internal representations mature or whether full network depth is required for stable outputs. To quantify depth utilization, we measure \textit{Effective Depth (ED)} following \cite{csordás2025languagemodelsusedepth}, using two complementary probes that capture internal feature evolution and output stabilization. Quantitative results across the three agentic domains are summarized in Table~\ref{tab:effective_depth}.

\parahead{Setting.}
For quantitative comparison, we replicate the effective-depth analysis across multiple model families, including Qwen3-30B-Instruct, Qwen3-30B-Thinking, GLM-4.5-Air, and Minimax-M2, over the three agentic domains studied in this section. 

\parahead{Effective Depth via Feature Orientation.}
Following \cite{csordás2025languagemodelsusedepth}, we define \textbf{Cosine Effective Depth} ($l_{cos}$) as the layer where residual cosine similarity makes its final sustained transition from negative to positive. This boundary separates the \textit{Construction Phase}, where representations are still being actively re-oriented, from the \textit{Refinement Phase}, where features are primarily amplified.

\parahead{Effective Depth via Output Stability.}
To estimate when intermediate representations become sufficient for final predictions, we apply \textbf{Logit Lens}~\cite{belrose2025elicitinglatentpredictionstransformers}. We report ED using two criteria:
(i) \textbf{Logit Lens KL}, defined as the layer at which the KL divergence between intermediate and final output distributions falls below half of its maximum; and
(ii) \textbf{Logit Lens Overlap}, defined as the layer at which the top-5 token overlap with the final distribution exceeds 0.3.

\begin{table*}[t]
    \centering
    \small 
    \setlength{\tabcolsep}{2.5pt} 
    
    \resizebox{\textwidth}{!}{
        \begin{tabular}{lcccccccccccccccccc}
            \toprule
            \multirow{3}{*}{Model} & \multicolumn{6}{c}{Cosine Similarity} & \multicolumn{6}{c}{Logit Lens KL} & \multicolumn{6}{c}{Logit Lens Overlap} \\
            \cmidrule(lr){2-7} \cmidrule(lr){8-13} \cmidrule(lr){14-19}
             
            & \multicolumn{2}{c}{\textbf{DeepRes.}} & \multicolumn{2}{c}{\textbf{Code}} & \multicolumn{2}{c}{\textbf{Tabular}} & \multicolumn{2}{c}{\textbf{DeepRes.}} & \multicolumn{2}{c}{\textbf{Code}} & \multicolumn{2}{c}{\textbf{Tabular}} & \multicolumn{2}{c}{\textbf{DeepRes.}} & \multicolumn{2}{c}{\textbf{Code}} & \multicolumn{2}{c}{\textbf{Tabular}} \\
           
            \cmidrule(lr){2-3} \cmidrule(lr){4-5} \cmidrule(lr){6-7} \cmidrule(lr){8-9} \cmidrule(lr){10-11} \cmidrule(lr){12-13} \cmidrule(lr){14-15} \cmidrule(lr){16-17} \cmidrule(lr){18-19}
             & ED & Ratio & ED & Ratio & ED & Ratio & ED & Ratio & ED & Ratio & 
    ED & Ratio & ED & Ratio & ED & Ratio & ED & Ratio \\
            \midrule
            
            Qwen3-Instruct & 19 & 0.40 & 20 & 0.42 & 19 & 0.40 & 44 & 0.92 & 44 & 0.92 & 45 & 0.94 & 45 & 0.94 & 45 & 0.94 & 46 & 0.96 \\
            
            Qwen3-Thinking & 19 & 0.40 & 20 & 
    0.42 & 20 & 0.42 & 44 & 0.92 & 44 & 0.92 & 45 & 0.94 & 45 & 0.94 & 45 & 0.94 & 47 & 0.98 \\
            
            \textbf{GLM-4.5-Air} & \textbf{13} & \textbf{0.28} & \textbf{34} & \textbf{0.74} & \textbf{34} & \textbf{0.74} & \textbf{31} & \textbf{0.67} & \textbf{32} & \textbf{0.70} & \textbf{36} & \textbf{0.78} & \textbf{36} & \textbf{0.78} & \textbf{39} & \textbf{0.85} & \textbf{41} & \textbf{0.89} \\
           
            Minimax-M2 & 16 & 0.26 & 29 & 0.47 & 20 & 0.32 & 55 & 0.89 & 56 & 0.90 & 58 & 0.94 & 62 & 1.00 & 62 & 1.00 & 62 & 1.00 \\
            \bottomrule
        \end{tabular}
    } 

\caption{
Effective depth (ED) and normalized depth ratio $(\mathrm{ED}+1)/L$ across model families and agentic domains, computed from Residual Cosine Similarity, Logit Lens KL, and Logit Lens Overlap. The table highlights systematic differences in how model families allocate depth between internal feature construction and output stabilization.
}
    \label{tab:effective_depth}
\end{table*}

\parahead{Comparative Analysis: Architectural Signatures of Depth.}
Table~\ref{tab:effective_depth} reveals two distinct structural patterns across model families.

\textit{\textbf{Construction-refinement gap (Qwen and Minimax).}} Qwen3 exhibits a highly consistent separation between internal feature construction and output convergence across all three domains. Its \textit{Cosine ED} remains shallow (ratio $0.40$-$0.42$), whereas both Logit Lens probes stabilize only near the end of the network (KL ratio $0.92$-$0.94$; Overlap ratio $0.94$-$0.98$), indicating that although active feature re-orientation completes relatively early, a large fraction of depth is still used to progressively refine already-formed representations before the final output stabilizes. Minimax-M2 shows the same qualitative construction-refinement gap, but with substantially greater domain variation: its \textit{Cosine ED} ranges from $0.26$ to $0.47$, while Logit-based ED remains consistently deep (KL ratio $0.89$-$0.94$; Overlap ratio $1.00$ across all three domains), suggesting that Minimax also devotes a large portion of depth to refinement, but does so with less stable feature-construction depth across tasks.

\textit{\textbf{Shared-expert divergence (GLM).}} GLM exhibits a distinctly different and strongly domain-dependent pattern. In Code Generation and Tabular Processing, its \textit{Cosine ED} reaches $0.74$, substantially deeper than in the other model families, while its KL-based Logit Lens ED remains close ($0.70$ and $0.78$), indicating that active feature re-orientation persists deep into the network even as output distributions are already approaching stability. Although the Overlap-based ED is somewhat deeper ($0.85$ and $0.89$), the gap remains much smaller than in Qwen3 or Minimax-M2. By contrast, in Deep Research, GLM falls back to a much shallower \textit{Cosine ED} ($0.28$), while its Logit-based ED remains moderate ($0.67$ for KL and $0.78$ for Overlap), suggesting that GLM allocates depth differently depending on task structure.

A definitive causal explanation would require explicit routing analysis. However, one plausible hypothesis is that GLM's hybrid \textit{Shared-Sparse Expert} architecture induces this domain dependence: specialized tasks such as code and tabular reasoning may recruit sparse experts for deeper feature transformation, whereas more open-ended research trajectories may rely more heavily on shared experts, leading to shallower construction dynamics and comparatively earlier stabilization.

\subsection{Further Discussion}

Our findings suggest several directions for future work. The progressive expansion of causal dependencies across turns indicates that agentic computation may be allocated conditionally on interaction state, rather than uniformly across depth. This raises the possibility of linking mechanistic depth analysis with adaptive-compute and routing-based inference mechanisms~\cite{heakl2025drllmdynamiclayerrouting,raposo2024mixtureofdepthsdynamicallyallocatingcompute,dehghani2019universaltransformers}.

The correction-dominant residual dynamics observed in later turns suggest that long-horizon agent behavior depends not only on feature accumulation, but also on repeated state revision. A natural next step is to study how these internal correction patterns interact with external memory, retrieval, and long-context management in agent systems~\cite{packer2024memgptllmsoperatingsystems,liu2023lostmiddlelanguagemodels}.

The architecture-dependent depth signatures observed across Qwen, Minimax, and GLM suggest that effective agent design may need to be architecture-aware. Some model families appear to rely on prolonged late-stage refinement, whereas others show more domain-dependent depth allocation, potentially reflecting differences in conditional computation and expert routing~\cite{shazeer2017outrageouslylargeneuralnetworks,fedus2022switchtransformersscalingtrillion,deepseekai2025deepseekv3technicalreport}. Future work could combine layer-wise tracing with explicit routing or expert-level analysis to clarify when deeper computation reflects semantic planning, memory repair, or output calibration~\cite{heakl2025drllmdynamiclayerrouting,raposo2024mixtureofdepthsdynamicallyallocatingcompute}.

\section{Conclusion}

In this work, we present a systematic mechanistic study of layer-wise dynamics in autonomous LLM agents during multi-turn planning. Moving beyond static benchmarks, we show that agentic reasoning induces a distinct pattern of adaptive depth allocation. As interaction trajectories unfold, models progressively recruit broader and deeper computation, while residual dynamics become increasingly correction-dominant in later turns. Quantitative probing further reveals a substantial construction-refinement gap: semantic direction often forms relatively early, yet deep layers remain necessary to stabilize final outputs. Across domains and model families, these results provide mechanistic evidence that sequential planning, tool use, and iterative state tracking place distinctive demands on architectural depth, underscoring the functional importance of deep models in autonomous agent settings.

\bibliographystyle{splncs04} 
\bibliography{samplepaper}

\end{document}